\newcommand{\bx}{\mathbf x}
\newcommand{\be}{\mathbf e}
\newcommand{\bg}{\mathbf g}

\newcommand{\bp}{\mathbf p}
\newcommand{\bq}{\mathbf q}
\newcommand{\bw}{\mathbf w}

\newcommand{\bs}{\mathbf s}
\newcommand{\bW}{\mathbf W}

\makeatletter
\DeclareRobustCommand\onedot{\futurelet\@let@token\@onedot}
\def\@onedot{\ifx\@let@token.\else.\null\fi\xspace}
\def\T{{\!\top}}
\def\eg{\emph{e.g}\onedot} 
\def\ie{\emph{i.e}\onedot}

\def\wrt{w.r.t\onedot}

\def\sexyname{SMA\xspace}

\documentclass[10pt,journal,compsoc]{IEEEtran}
\ifCLASSOPTIONcompsoc
  \usepackage[nocompress]{cite}
  \usepackage{graphicx}
  \usepackage{comment}
  \usepackage{amsmath,amssymb} 
  \usepackage{color}
  \usepackage{epsfig}
  \usepackage{multirow}
  \usepackage[marginal]{footmisc}
  \usepackage{booktabs}
  \usepackage{float}
  \usepackage{caption}
  \usepackage{hyperref}
  \hypersetup{colorlinks=true,urlcolor=blue}
  \usepackage{xspace}
  \usepackage{bbm}
  \usepackage{dsfont}
\else
  \usepackage{cite}
\fi

\begin{document}
%
\title{Structured Multimodal Attentions for TextVQA}
%
%
%
%

\author{Chenyu Gao, Qi Zhu, Peng Wang$^\dagger$, Hui Li, Yuliang Liu, Anton van den Hengel, Qi Wu$^\dagger$
\IEEEcompsocitemizethanks{\IEEEcompsocthanksitem $^\dagger$PW and QW are corresponding authors.
\IEEEcompsocthanksitem C. Gao, Q. Zhu and P. Wang are with the Northwestern Polytechnical University. H. Li, Y. Liu, A. van den Hengel and Q. Wu are with the University of Adelaide}
\thanks{Manuscript received April 19, 2005; revised August 26, 2015.}}

%
%

\markboth{Journal of \LaTeX\ Class Files,~Vol.~14, No.~8, August~2020}%
{Gao \MakeLowercase{\textit{et al.}}: Bare Demo of IEEEtran.cls for Computer Society Journals}
%



\IEEEtitleabstractindextext{%
\begin{abstract}
Text based Visual Question Answering (TextVQA) is a recently raised challenge requiring models to read text in images and answer natural language questions by jointly reasoning over the question, textual information and visual content. Introduction of this new modality - Optical Character Recognition (OCR) tokens ushers in demanding reasoning requirements. Most of the state-of-the-art (SoTA) VQA methods fail when answer these questions because of three reasons: (1) poor text reading ability; (2) lack of textual-visual reasoning capacity; and (3) choosing discriminative answering mechanism over generative couterpart (although this has been further addressed by M4C). 
In this paper, we propose an end-to-end structured multimodal attention (SMA) neural network to mainly solve the first two issues above. SMA first uses a structural graph representation to encode the object-object, object-text and text-text relationships appearing in the image, and then designs a multimodal graph attention network to reason over it. Finally, the outputs from the above modules are processed by a global-local attentional answering module to produce an answer splicing together tokens from both OCR and general vocabulary iteratively by following M4C. 
Our proposed model outperforms the SoTA models on TextVQA dataset and two tasks of ST-VQA dataset among all models except pre-training based TAP. Demonstrating strong reasoning ability, it also won first place in TextVQA Challenge 2020. 
We extensively test different OCR methods on several reasoning models and investigate the impact of gradually increased OCR performance on TextVQA benchmark. With better OCR results, different models share dramatic improvement over the VQA accuracy, but our model benefits most blessed by strong textual-visual reasoning ability. 
To grant our method an upper bound and make a fair testing base available for further works, we also provide human-annotated ground-truth OCR annotations for the TextVQA dataset, which were not given in the original release. 
The code and ground-truth OCR annotations for the TextVQA dataset are available at https://github.com/ChenyuGAO-CS/SMA
\end{abstract}

}

\maketitle

\IEEEdisplaynontitleabstractindextext

%
\IEEEpeerreviewmaketitle

\section{Introduction}
\IEEEPARstart{V} \,{isual} Question Answering (VQA)~\cite{antol2015vqa} has shown great progress by virtue of the development of deep neural networks. However, recent studies~\cite{STVQA,vizwiz,TexVQA} show that most VQA models fail unfortunately on a type of questions requiring understanding the text in the image. The VizWiz~\cite{vizwiz} firstly identified this problem and found that nearly a quarter of questions asked by visually-impaired people are text-reading related. Singh~\cite{TexVQA} systematically studied this problem and introduced a novel dataset TextVQA that only contains questions requiring the model to read and reason about the text in the image. 

\begin{figure}[t]
\begin{center}
   \includegraphics[width=0.49\textwidth]{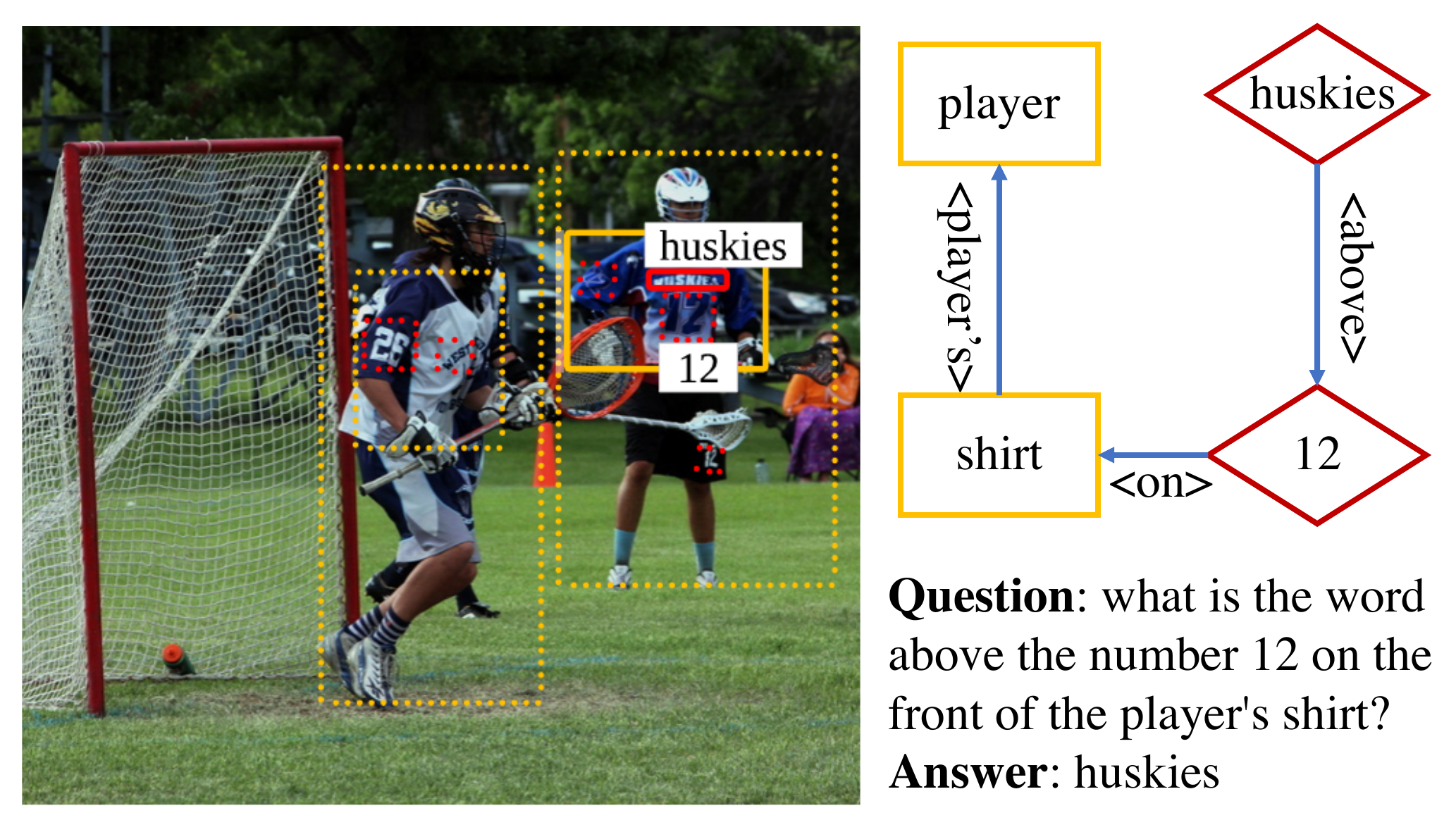}
\end{center}
   \caption{
    {\bf (Left)} In Question Conditioned Graph Attention Module, we construct a heterogeneous graph. We illustrate objects in yellow and OCR tokens in red. While unbroken-line boxes represent nodes most relevant to the question, dashed-line ones are other nodes. Understanding multiple relationships is crucial to answer this question, \eg ``word above the number 12'' is a text-text relation. {\bf (Right)} Abstract entities and relationships. Red diamonds are OCR tokens and yellow rectangles are objects. Blue arrows are relationships.
   }
\label{fig:title}
\end{figure}

Three key abilities that are essential in tackling the TextVQA problem are \textit{reading}, \textit{reasoning} and \textit{answering}, which are also the main reasons why state-of-the-art (SoTA) VQA models turn out badly on this task. The \textit{reading} ability relies on the Optical Character Recognition (OCR) techniques to detect and recognize the text appeared in the image accurately which have already been a long-standing sub-field of computer vision, and the \textit{reasoning} needs a model to jointly reason over the visual content and OCR text in the image. 
The SoTA VQA models~\cite{gao2018motion,yu2019deep} may gain strong reasoning abilities on visual content and natural language questions via some sophisticated mechanisms such as attention~\cite{bahdanau2014neural} and memory networks~\cite{weston2014memory}, but none of them can read the ``text'' in images accurately, not to mention reasoning over them. LoRRA \cite{TexVQA}, the method provided in TextVQA, although equipping an OCR model to read text, has results that are not outstanding due to a lack of deep reasoning among text and visual content. 
As to the \textit{answering} aspect, almost all of the SoTA VQA models choose to use a discriminative answering module because it is easy to be optimized and leads to better performance on traditional VQA datasets. However, the answer in TextVQA is normally a combination of detected OCR tokens from the image and general text tokens, thus the answer vocabulary is not fixed. The discriminative answering module may limit the output variety. 

Figure \ref{fig:title} shows an example from TextVQA that involves several types of relationships. For instance, ``the front of shirt'', ``player's shirt'' are object-object links; ``word printed on the front of the player's shirt'' is a text-object bond and the ``word above the number 12'' is a text-text relation.
In this paper, to enhance the relationship reasoning ability, we introduce an \sexyname model to reason over a graph that has multiple types of relationships. 
Specifically, a question self-attention module firstly decomposes questions into six sub-components that indicate \textit{objects, object-object relations, object-text relations, texts, text-text relations} and \textit{text-object relations}. 
A role-aware graph is then constructed with objects/texts as nodes. The connections between nodes are decided by the relative distance. Then the graph is updated by using a question conditioned graph attention module. In this model, instead of using the whole question chunk to guide the graph updating, only certain types of question components extracted from the question self-attention module can be used to update the corresponding graph components. For example, \textit{object} related question feature is for object nodes and the \textit{object-text} related question feature is only for the object-text edge updating process. 
Finally, to solve the aforementioned \textit{answering} issue, we introduce the iterative answer prediction mechanism in M4C~\cite{hu2019iterative} into our global-local attentional module but we replace the first step input $\mathsf{<begin>}$ with the summarized global features of question, object and text, together with local OCR embeddings. The original input of first decoding step of M4C is a special token $\mathsf{<begin>}$ without more information, while the summarized global features of us include comprehensive information of our question conditioned graph, which bring an improvement of nearly $0.5\%$ to our model. 
All of the features used in the middle process are automatically learnt in an end-to-end manner, which allows us to flexibly adapt to different instances with different types of relations. The only supervision is the Ground-Truth answer of each instance.

We verify the effectiveness of our proposed \sexyname model on recently released TextVQA \cite{TexVQA} and ST-VQA \cite{STVQA} datasets and we outperform the previous SoTA models on TextVQA and the first two tasks of ST-VQA. Our proposed \sexyname model also won the TextVQA Challenge 2020\footnote{https://evalai.cloudcv.org/web/challenges/challenge-page/551/leaderboard/1575}.

To further study the contribution of ``\textit{reading}'' part and ``\textit{reasoning}'' part, we investigate how much the TextVQA accuracy will be affected by the OCR performance if a fixed reasoning model is used. 
We conduct text detection and recognition on TextVQA using SoTA text detector and recognizer. In comparison with the results from Rosetta OCR~\cite{borisyuk2018rosetta}, it is found that all our models enjoy an improvement, as well as the LoRRA and M4C~\cite{hu2019iterative}, but our \sexyname benefits more with better OCR result, which proves that our model has better textual-visual reasoning ability. 
To completely peel off the impact of OCR for investigating the real reasoning ability, we ask AMT workers to annotate all the text appeared in the TextVQA dataset, which leads to $709,598$ ground-truth OCR annotations. These annotations were not given in the original TextVQA and we will release them to the community for a fair comparison. We also report the performance of LoRRA, M4C and our best model by giving the ground-truth OCR, in order to test solely the reasoning ability of the model. A new upper bound is also given by using the ground-truth OCR annotations.

In summary, our contributions are threefold:
\begin{enumerate}
  \item  We propose a structured multimodal attentional (\sexyname) model that can effectively reason over structural text-object graphs and produce answers in a generative way.  Thanks to the adopted graph reasoning strategy, the proposed model achieves better interpretability.
  \item We study the contribution of OCR in the TextVQA problem and provide human-annotated ground-truth OCR labels to complete the original TextVQA dataset. This allows followers in the community to only evaluate their models' \textit{reasoning} ability, under a perfect \textit{reading} situation.
  \item Our \sexyname model outperforms existing state-of-the-art TextVQA models on both TextVQA dataset and two tasks of ST-VQA dataset (except a pre-training based model TAP~\cite{yang2020tap}, which is not directly comparable), leading to a champion model on TextVQA Challenge 2020.
\end{enumerate}

\section{Related Work}

\subsection{Text based VQA}
Straddling the field of computer vision and natural language processing, Visual Question Answering (VQA) has attracted increasing interests since the release of large-scale VQA dataset~\cite{antol2015vqa}. A large number of methods and datasets emerged: VQA datasets such as CLEVR~\cite{johnson2017clevr} and FigureQA~\cite{FigureQA} have been introduced to study visual reasoning purely, without the consideration of OCR tokens; Wang \textit{et al.}~\cite{FVQA} introduced a dataset that explicitly requires external knowledge to answer a question.

Reading and reasoning over text involved in an image are of great value for visual understanding, since text contains rich semantic information which is the main concern of VQA. Several datasets and baseline methods are introduced in recent years aiming at the study of joint reasoning ability over visual and text contents. For example, Textbook QA~\cite{TextbookQA} asks multimodal questions given text, diagrams and images from middle school textbooks.   
FigureQA~\cite{FigureQA} needs to answer questions based on synthetic scientific-style figures like line plots, bar graphs or pie charts. 
DVQA~\cite{DVQA} assesses bar-chart understanding ability in VQA framework.  
In these datasets, texts are machine-printed and appear in standard font with good quality, which alleviate the challenging text recognition work.  
Vizwiz~\cite{vizwiz} is the first dataset that requires text information for question answering, given images captured in natural scenes. 
Nevertheless, $58\%$ of the questions are ``unanswerable'' because of the poor image quality, which makes the dataset inappropriate for training an effective VQA model and studying the problem systematically.

Most recently, TextVQA~\cite{TexVQA} and ST-VQA~\cite{STVQA} are proposed concurrently to highlight the importance of text reading in images from natural scene in the VQA process. 
LoRRA was proposed in TextVQA which uses a simple Updn~\cite{bottomuptopdown} attention framework on both image objects and OCR texts for inferring answers. The model was then improved by using a BERT~\cite{BERT} based word embedding and a Multimodal Factorized High-order pooling based feature fusion method, and achieved the winner in TextVQA challenge. Compared to TextVQA where any question is allowed once text reading is required, all questions in ST-VQA can be answered unambiguously directly by text in images. Stacked Attention Network (SAN)~\cite{SAN} is adopted in ST-VQA as a baseline, by simply concatenating text features with image features for answer classification. The answering modules in previous models such as LoRRA~\cite{TexVQA} encounters two bottlenecks. One serious setback is that they view dynamic OCR space as invariant indexes, and the other is the disability to generate long answers composed of more than one word. MM-GNN~\cite{gao2020multi} focused on construct visual graph, semantic graph and numeric graph for representing the information of scene texts. M4C~\cite{hu2019iterative} firstly tackles both problems by a transformer decoder and a dynamic pointer network. Following on M4C, the LaAP~\cite{han2020finding} further predicts the OCR position as the evidence for final predicted answer, SA-M4C~\cite{sam4c} replaced the original self-attention layers with their novel spatially aware self-attention layers, 
Singh \textit{et al.}~\cite{singh2021textocr} proposed an OCR dataset (900k annotated arbitrary-shaped words) with an end-to-end PixelM4C model, which connected Mask TextSpot- ter (MTS) v3~\cite{liao2020mask} with M4C~\cite{hu2019iterative} and can extract OCR and do text-based tasks at the same time. 
MTXNet~\cite{rao2021first} focused on textual and visual explanations, which proposed a dataset TextVQA-X for producing explanations and found that existing TextVQA models can be adapted to generate multimodal explanations easily. 
Yong \textit{et al.}~\cite{yang2020tap} are aming at proposing pre-training tasks to learn a better aligned representation of of text word, visual object, and scene text and make performance improvement on text based tasks, such as TextVQA and Textcaps~\cite{sidorov2020textcaps}. 
In this work, we focus on explicitly modeling relationships between objects, texts and object-text pairs, and achieved better performance and interpretability than previous TextVQA approaches.

\subsection{Graph Networks in Vision and Language}
Graph networks have received a lot of attention due to their expressive power on structural feature learning. They can not only capture the node features themselves, but also encode the neighbourhood properties between nodes in graphs, which is essential for VQA and other vision-and-language tasks that need to incorporate structures in both spatial and semantic information. For instance, Teney~\emph{et al.}~\cite{Teney2017} construct graphs over image scene objects and over question words respectively to exploit the structural information in these representations. The model shows significant improvements in general VQA tasks. Narasimhan~\emph{et al.}~\cite{Narasimhan2018} perform finer relation exploration for factual-VQA task~\cite{FVQA} by taking into account a list of facts via Graph Convolution Networks (GCN) for correct answer selection. 
The work \cite{Will2018} learns a question specific graph representation for input image in VQA, capturing object interactions with the relevant neighbours via spatial graph convolutions. MUREL~\cite{MUREL} goes one step further to model spatial-semantic pairwise relations between all pairs of regions for relation reasoning, in addition to a rich vectorial representation for interaction between region's visual content and question. 

Our work also uses graph as the representation, but different from previous methods that use a fully-connected graph to connect all the objects, our task needs to take into account both visual elements and text information from image, which are essentially heterogeneous. A role-aware graph is constructed that considers different roles of nodes (such as object and text) and edges (``object-object'', ``text-text'' and ``object-text''), which results in a much better cross-modality feature representation for answer inferring.

\subsection{Scene Text Detection and Recognition}
The result of OCR (including both text detection and recognition) in natural scene images plays an important role in TextVQA. However, due to the extreme diversity of text patterns and the highly complicated background, OCR in natural scenes by itself is a really challenging task, and attracts much attention in the computer vision community. Most works address it separately by designing high performance text detector~\cite{Yuliang19,Lyucvpr18,jiaya19,PSENet19} and sophisticated word recognizer~\cite{Cheng2018AON,li2019aaai,shiCVPR2016,textTrans,ESIR19}, which can be combined together to obtain text contents in an image finally. Some recent works are also proposed to address both text detection and recognition in one framework~\cite{hetong2018,li2017towards,FOTS2018,Maskspot2018}.
Driven by deep neural networks and large scale datasets, substantial progress has been made in both performance and speed. Methods are improved from only handling horizontal text to tackling text with arbitrary shapes. The great improvement sets the stage for VQA based on both visual information and text clues. Rosetta OCR~\cite{borisyuk2018rosetta} was adopted in LoRRA for text spotting, which is a two-step model with Faster-RCNN based framework for text detection and a fully convolutional model with CTC loss for word recognition. Irregular text (oriented, distorted) cannot be read well by Rosetta. In our model, an SoTA Sequential-free Box Discretization (SBD) model~\cite{SBD} is used firstly for scene text detection and a robust transformer based network~\cite{textTrans} is employed for word recognition (denoted as ``SBD-Trans OCR''), which achieves a better OCR result and leads to an improvement on TextVQA task. 

\begin{figure}[t]
\begin{center}
   \includegraphics[width=0.48\textwidth]{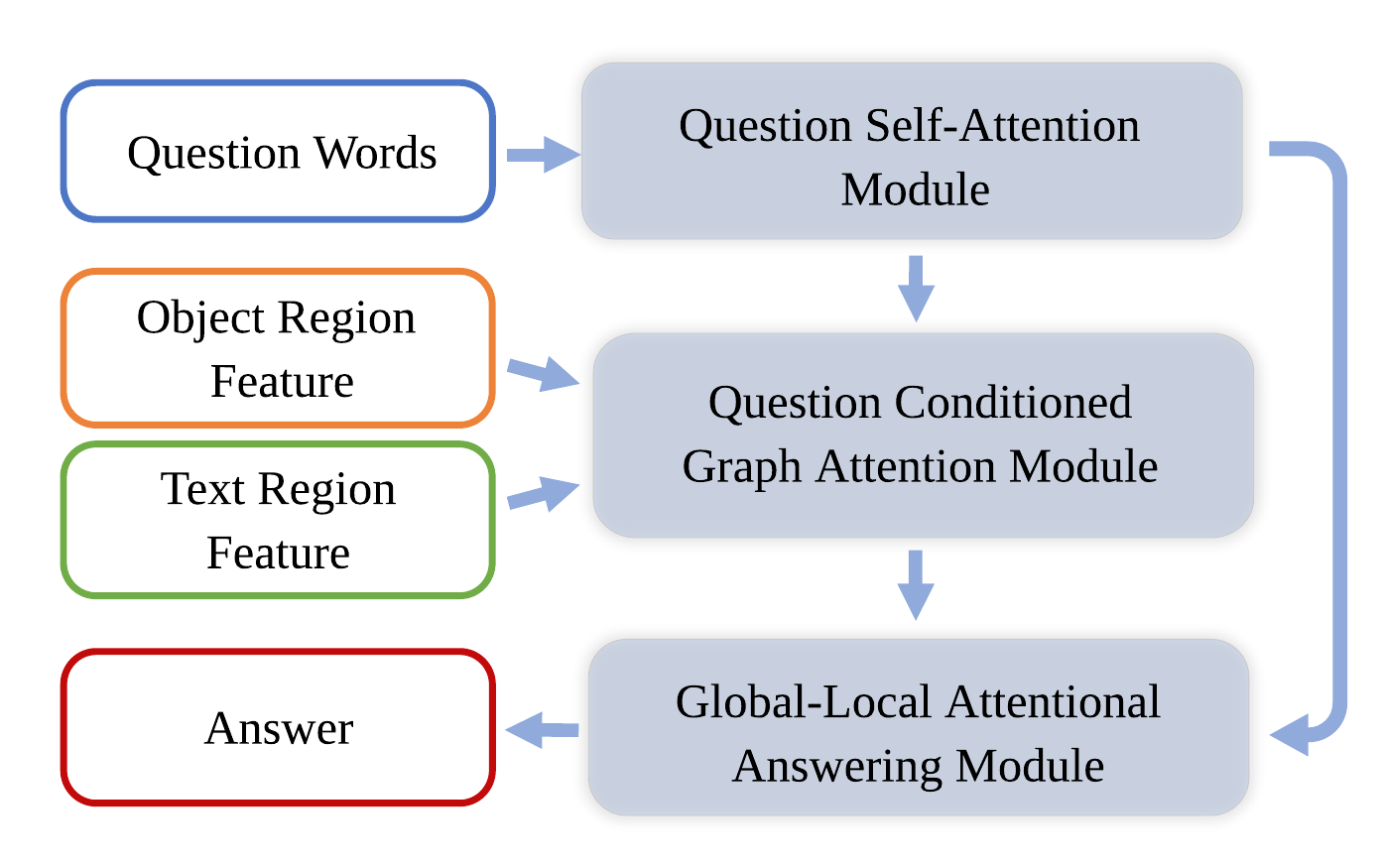}
\end{center}
   \caption{
   Three modules in our SMA model. Question Self-Attention Module decomposes the question into guiding signals that guide Graph Attention Module to reason over a graph, and a Global-Local Attentional Answering module to generate an answer.
   }
\label{fig:fullmodel}
\end{figure}

\begin{figure*}[t]
	\centering
	\begin{center}
	\includegraphics[width=0.98\textwidth]{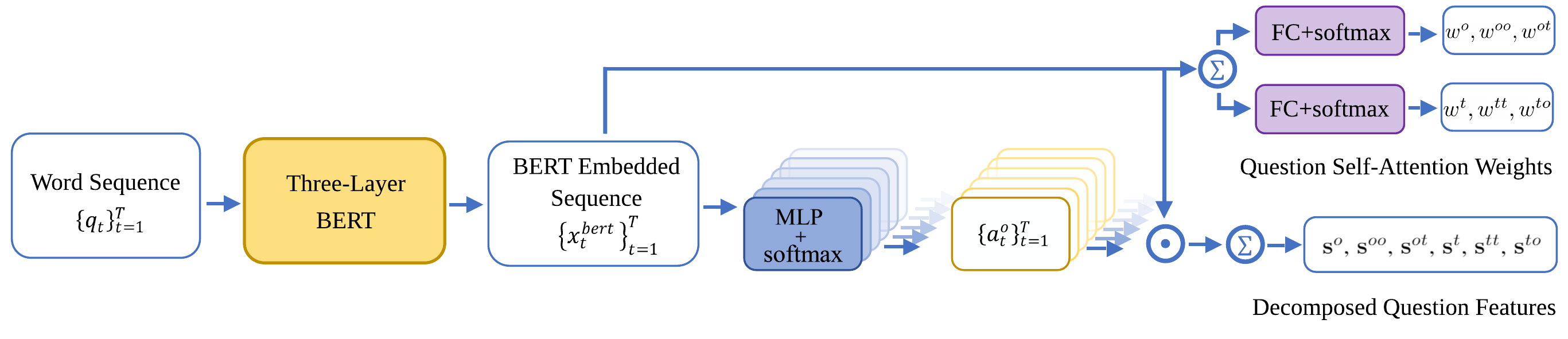}
	\caption{An overview of Question Self-Attention Module. Input word sequence of a question, we get two kinds of attention weights: question self-attention weights which account for prior probability in the whole graph and fine-grained decomposed question features for corresponding nodes or edges. 
	}
	\label{fig:languageSelfAttention}
	\end{center}
\end{figure*}
\section{Method}

In this section we introduce our Structured Multimodal Attentions (\sexyname) model. 
At a high level, \sexyname is composed of three modules, as shown in Figure~\ref{fig:fullmodel}: 
(1) a question self-attention module that decomposes questions into six sub-components \wrt different roles in our constructed object-text graph.
(2) a question conditioned graph attention module that reasons over the graph under the guidance of the above question representations and infers the importance of different nodes as well as their relationships. 
and 
(3) a global-local attentional answering module which can generate answers with multiple words stitching together. 
Our model is an end-to-end framework and the only supervision is the Ground-Truth answer of each instance. All of the sub-features are automatically learnt during the training process, which allows us to flexibly adapt to different instances with different types of relations. We detail each module in the following sections.

\noindent {\bf Notation} In the remainder of this paper, matrices are denoted by bold capital letters and column vectors are denoted by bold lower-case letters. $\circ$ represents element-wise product. 
$[;]$ refers to concatenation.

\subsection{Question Self-Attention Module}
\label{sec:lan}
Since a TextVQA question may include not only information of object and text nodes, but also four categories of relationships between them (object-object, object-text, text-text and text-object), our question self-attention module (see Figure~\ref{fig:languageSelfAttention}) firstly divides a question into six sub-components. It is similar to domain knowledge guided multi-head attention, as different question components correspond to different attention heads and they are fine-grained and specifically applied to every single node and edge. Although this is inspired by~\cite{andreas2016neural,yu2018mattnet}, our modules are more fine-grained and are carefully designed for the TextVQA task. 

Formally, given a question $Q$ with $T$ words $q = \{q_t\}_{t=1}^T$, $\{\bx^{bert}_t\}_{t=1}^T$ is obtained by using pre-trained BERT~\cite{BERT}. The decomposed question features ($\bs^{o}$, $\bs^{oo}$, $\bs^{ot}$, $\bs^{t}$, $\bs^{tt}$, $\bs^{to}$) are considered as question representations decomposed \wrt object nodes ({\bf o}), object-object ({\bf oo}) edges, object-text ({\bf ot}) edges, text nodes ({\bf t}), text-text ({\bf tt}) edges and text-object ({\bf to}) edges. 
Taking $\bs^{o}$ as example, the computation is performed as follows:

\begin{equation}
	\begin{split}
	&{a_t^o = \mathbf{softmax}(\mathrm{MLP}^a_{obj}(\bx^{bert}_t)), \quad t = 1, \dots, T, }\\
	&\bs^o = \textstyle{\sum_{t=1}^{T}} a_t^o  \bx^{bert}_t.
	\end{split}
	\label{selfatttt}
\end{equation}

Other features are computed in a same way and these decomposed question features are used as guiding signals when performing question conditioned graph attention in Section~\ref{sec:lgGraphAttention}. 

We also apply another series of transformations on the averaged hidden states, in order to generate two sets of question self-attention weights which will be used for final feature combination as prior probability in Section \ref{sec:ans_mod}.
To be more specific, $\{ w^{o}, w^{oo}, w^{ot} \}$ and $\{ w^{t}, w^{tt}, w^{to} \}$ are used to generate object feature $\bg_{obj}$ and text feature $\bg_{text}$ respectively. 

All of the above mentioned sub-feature are learnt in an end-to-end training manner without any hand-coded supervision for simple design and high adaptability. The attention weights are learnt in language self-attention module, what kind of role they will attain depend on the type of query key they queried (in our case, keys are nodes' and edges' representations). 

\begin{figure*}[t]
	\centering
	\begin{center}
	\includegraphics[width=\textwidth]{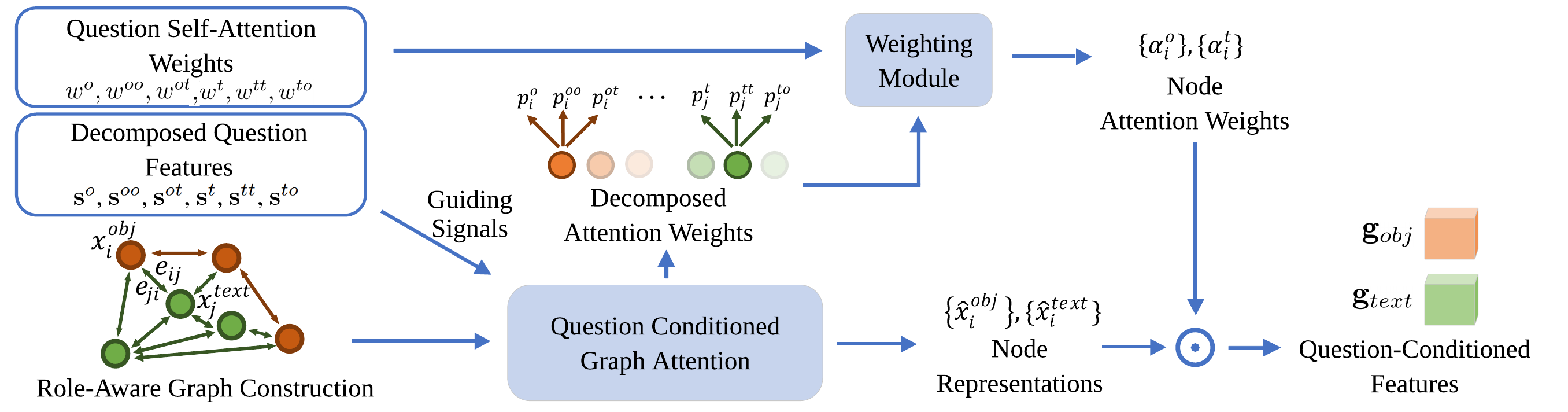}
	\caption{An overview of Question Conditioned Graph Attention Module. This module builds a heterogeneous graph whose mixed nodes are shown in different colors. Guiding signals help produce attention weights, fused which with node representations we get question-conditioned features.
	}
	\label{fig:QCGA}
	\end{center}
\end{figure*}

\subsection{Question Conditioned Graph Attention Module.}
\label{sec:lgGraphAttention}
The question conditioned graph attention module (as shown in Figure \ref{fig:QCGA}) is the core of our network, which generates a heterogeneous graph over both objects and texts of an image and then reasons over it. 

\begin{figure}[t!]
	\begin{center}
		\includegraphics[width=0.47\textwidth]{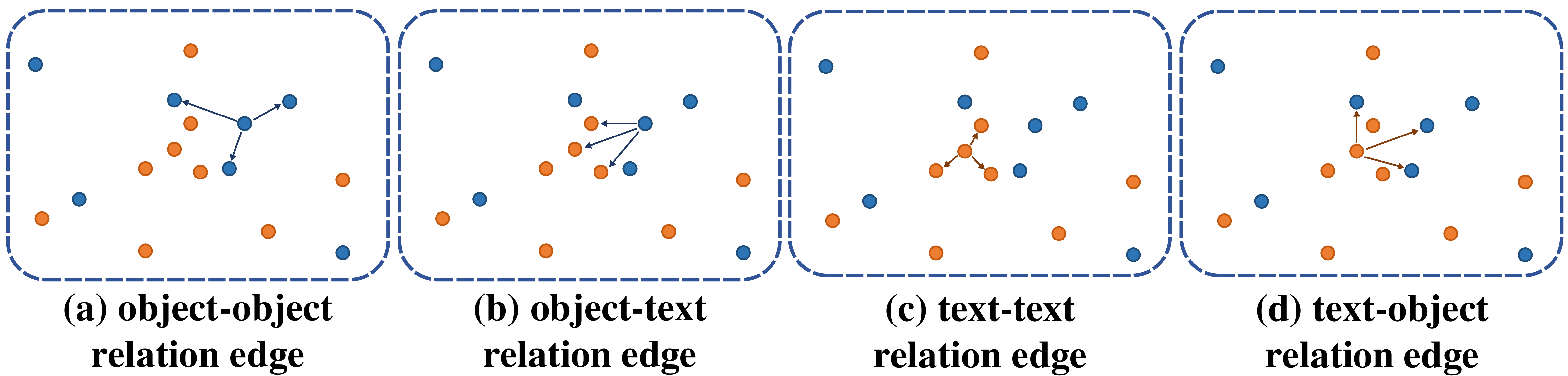}
	\end{center}
	\caption{ Schematic diagram of four kinds of edge structures. Blue dots represent object nodes, and orange dots represent text nodes. Here we set $k=3$ for simplicity, a node will only have edges with its $k$-nearest neighbours.}
	\label{fig:edgeBuild}
\end{figure}

\noindent{\bf Role-aware Heterogeneous Graph Construction.}
\label{sec:role}
`Role' denotes different type of nodes. 
We construct a role-aware heterogeneous graph $\mathcal{G} = \{\mathcal{O}, \mathcal{T}, \mathcal{E}\}$ over object nodes and text nodes of an image $I$, 
where $\mathcal{O} = \{o_i\}_{i=1}^N$ is the set of $N$ object nodes, 
$\mathcal{T} = \{t_i\}_{i=N+1}^{N+M}$ is the set of $M$ text nodes and $\mathcal{E} = \{e_{ij}\}$ is the edge set. 
In our graph, 
an edge denotes the relationship between two particular nodes and 
each node can be connected to $k=5$ object nodes plus $k=5$ text nodes. 
It is apparent that nodes and edges in our graph have different roles, thus we call it a heterogeneous graph. `Role-awareness' means we explicitly use the role information of each node to construct the graph. 
We can further divide the edges into four sets according to their different roles: $\mathcal{E}^{oo}$ for {\bf oo} edges, $\mathcal{E}^{ot}$ for {\bf ot} edges, $\mathcal{E}^{tt}$ for {\bf tt} edges and $\mathcal{E}^{to}$ for {\bf to} edges. See Figure \ref{fig:edgeBuild} for an illustration. 


We then build the edge representation between two nodes based on their relative spatial relationship. Again here we build an {\bf oo} edge as an example. 
Suppose the center coordinate, width and height of a node $o_i$ are represented as $[x^c_{i}, y^c_{i}, w_i, h_i]$, and the top-left coordinate, bottom-right coordinate, width and height of another node $o_j$ are represented as $[{x^{tl}_j}, {y^{tl}_j}, {x^{br}_j}, {y^{br}_j}, {w_j}, {h_j}]$, 
then the associated edge representation is defined as 
$\be_{ij} = [\frac{x^{tl}_j-x^{c}_i}{w_i}, \frac{y^{tl}_j-y^{c}_i}{h_i}, \frac{x^{br}_j-x^{c}_i}{w_i}, \frac{y^{br}_j-y^{c}_i}{h_i},  \frac{w_j\cdot h_j}{w_i\cdot h_i}]$.

\noindent{\bf Question Conditioned Graph Attention.}
\label{sec:lan_graph_attn}
We use the decomposed question features $\bs$ in Section \ref{sec:lan} to reason on our role-aware graph constructed in the last section. 

We formulate the reasoning process as an attention mechanism, but instead of applying a global attention weights with single question features, we update different parts of the graph with different question features according to their roles. For example, the object-related question representation $\bs^o$ is used to guide the attention weights over object nodes, and $\bs^{to}$ is used to guide the text-object edge attention weights. Considering that there are six roles in the graph, 
we compute the attention weights respectively for object nodes ($\bp^{o}$), text nodes ($\bp^{t}$), object-object edges ($\bp^{oo}$), object-text edges ($\bp^{ot}$),
text-text edges ($\bp^{tt}$) and text-object edges ($\bp^{to}$).
The mechanism can be formulated as:
\begin{equation}
 \bp^{m} = \mathrm{Att}_{m}(\{\bx^{{obj}}\}, \{\bx^{{text}}\}, \{\be_{ij}\}, \bs^{m}),
\end{equation}
where $\mathrm{Att}_m$ is the attention mechanism to compute attention weights using question features and specific nodes/edges in graph, and $m=\{o,oo,ot,t,tt,to\}$. $\bx^{obj}$ and $\bx^{text}$ represent features extracted from isolated object and text regions respectively, which are then fed into the graph attention module to generate question-conditioned features. Now we describe how the $\mathrm{Att}_m$ is calculated based on different types of attention.

{\em 1) The node representation.}
An object node is represented by it's $2048$D appearance feature from a Faster R-CNN detector and $4$D bounding box feature with object’s relative bounding box coordinates $[\frac{x^{tl}_i}{W}, \frac{y^{tl}_i}{H}, \frac{x^{br}_i}{W}, \frac{y^{br}_i}{H}]$, where $W$ and $H$ represent the width and height of the image. Given the appearance features ${\{\bx_{fr,i}^{o}\}}_{i=1}^N$, 
and bounding box feature ${\{\bx_{bbox,i}^{o}\}}_{i=1}^N$ of a object, the representation of object node is calculated by:
\begin{equation}
{
	    \hat{\bx}_i^{obj} = \mathrm{LN}(\bW_{fr}^{o}\bx_{fr,i}^{o}) + \mathrm{LN}(\bW_{b}^{o}\bx_{bbox,i}^{o}),
	\label{attn_o}}
\end{equation}
where $\mathrm{LN}(\circ)$ is layer normalization; $\bW_{fr}^{o}$ and $\bW_{b}^{o}$ are linear transformation parameters to be learned. 

For text nodes, we also employ a combination of multiple features (referred to as Multi-Feats) to enrich OCR regions' representation as in~\cite{hu2019iterative}: 1) a $300$D FastText feature ${\{\bx_{ft,i}^{t}}\}_{i=N+1}^{N+M}$ is generated from a pre-trained FastText~\cite{FastText} embeddings, 2) a $2048$D appearance feature ${\{\bx_{fr,i}^{t}}\}_{i=N+1}^{N+M}$ is generated from the same Faster R-CNN detector as object nodes, 3) a $604$D Pyramidal Histogram of Characters (PHOC)~\cite{almazan2014word} feature ${\{\bx_{p,i}^{t}}\}_{i=N+1}^{N+M}$ and 4) a $4$D bounding box feature ${\{\bx_{bbox,i}^{t}}\}_{i=N+1}^{N+M}$. 
In addition to multi-feats, we also introduce a $512$D CNN feature ${\{\bx_{rec,i}^{t}}\}_{i=N+1}^{N+M}$ (referred to as RecogCNN), which is extracted from a transformer-based text recognition network~\cite{textTrans}.
The representation for text node are calculated by:
\begin{equation}
{
	\begin{aligned}
	    & \bx_{i}^{m} = \bW_{ft}^{t}\bx_{ft,i}^{t} + \bW_{fr}^{t}\bx_{fr,i}^{t} + \bW_{p}^{t}\bx_{p,i}^{t} + \bW_{rec}^{t}\bx_{rec,i}^{t}, \\
	    &\hat{\bx}_i^{text} = \mathrm{LN}(\bx_{i}^{m}) 
	    + \mathrm{LN}(\bW_{b}^{t}\bx_{bbox,i}^{t}),
	\end{aligned}
	\label{attn_o}}
\end{equation}
where $\bW_{ft}^{t}$, $\bW_{fr}^{t}$, $\bW_{p}^{t}$, $\bW_{rec}^{t}$ and $\bW_{b}^{t}$ are linear transformation parameters to be learned.

{\em 2) The node attention weights.} Due to the similarity of the calculation process of object node attention weights and text node attention weights, we choose $\bp^{o}$ for illustration:
Given the representation $\{\hat{\bx}_i^{obj}\}_{i=1}^N$ of an object node, 
the attention weights for object node is calculated under the guidance of $\bs^{o}$: 
\begin{equation}
{
	\begin{aligned}
		&p_i^{o'} = \bw_o^\T[\mathrm{ReLU}(\bW_{s}^{o} \bs^{o})\circ \mathrm{ReLU}(\bW_{x}^{o}\hat{\bx}_i^{obj})],\\
		&{p_i^{o} = \mathbf{softmax}(p_i^{o'}), \quad i = 1, \dots, N,}
	\end{aligned}
	\label{attn_o}}
\end{equation}
where $\bW_{s}^{o}$, $\bW_{x}^{o}$ and ${\bw_o}$ are linear transformation parameters to be learned.

For the attention weights of text nodes ($\bp^{t}$), we perform the same calculation but with independent parameters.

{\em 3) The edge attention weights.} The edge attention weights need to consider the relationship between two nodes. Because the calculation process of attention weights for different edge types $\bp^{oo}$, $\bp^{ot}$, $\bp^{tt}$ and $\bp^{to}$ are similar, we only show how $\bp^{oo}$ is computed.

\begin{figure*}[t]
	\centering
	\begin{center}
	\includegraphics[width=0.98\textwidth]{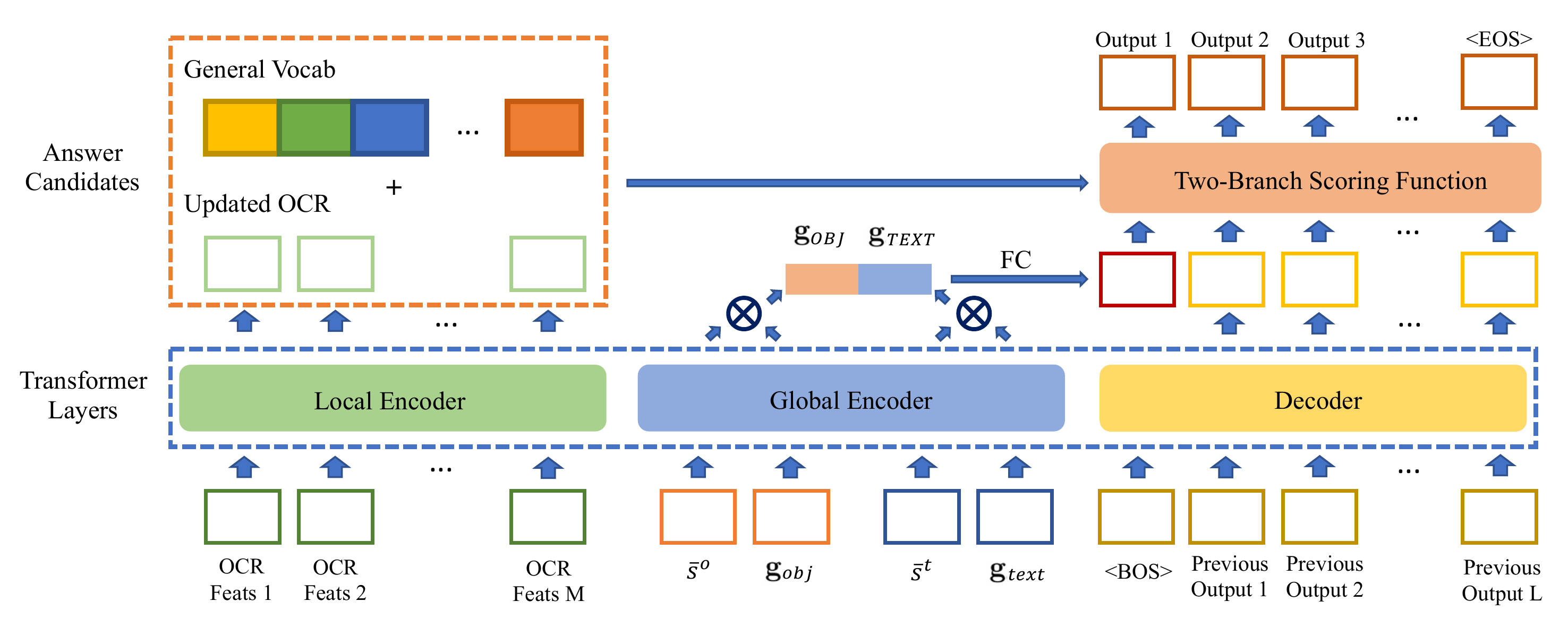}
	\caption{An overview of Global-Local Attentional Answering Module. The same transformer layers are split into three functionally different parts. Local Encoder updates local ocr embeddings. The results of Global Encoder are used to predict answer for the first timestep. General vocabulary and updated OCR consist answer candidates, from with we use a two-branch scoring function to select the answer in each timestep.
	}
	\label{fig:answerModule}
	\end{center}
\end{figure*}

There are mainly two steps.
Firstly, for each node $o_i$, 
we compute the attention weights $\boldsymbol{\bq}^{oo}_i = \{ q^{oo}_{ij} \}_{j \in \mathcal{N}^{oo}_i}$ over all the {\bf oo} edges connected to $o_i$:
\begin{equation}\label{attn_oo}
{
    \begin{aligned}
    &\hat{\bx}_{ij}^{oo} = \mathrm{MLP}([\be_{ij}; \hat{\bx}_i^{obj}]), \\
    &q_{ij}^{oo'} = {\bw_{oo}^\T}[\mathrm{ReLU}(\bW^{oo}_s\, \bs^{oo})\circ \mathrm{ReLU}(\bW^{oo}_x \, \hat{\bx}_{ij}^{oo})],\\
    &{q_{ij}^{oo} = \mathbf{softmax}( q^{oo'}_{ij}), \quad {j \in \mathcal{N}^{oo}_i},}
    \end{aligned}
}
\end{equation}
where 
$\bW_s^{oo}$ and $\bW_x^{oo}$ respectively maps the {\bf oo} edge related question representation $\bs^{oo}$ 
and the embedded edge features $\hat{\bx}_{ij}^{oo}$ into vectors of the same dimension. 
The attention weights ${\bq}_{i}^{oo}$ are normalized over neighborhood $\mathcal{N}_i^{oo}$ of $o_i$ via a softmax layer. 

At the second step, we calculate {\bf oo} edge attention weights $\bp^{oo} = \{ p_i^{oo} \}_{i=1}^{N}$ over all object nodes: 
\begin{equation}\label{attn_oo}
{
    \begin{aligned}
    &\tilde{\bx}_i^{oo} = \textstyle{\sum_{j \in \mathcal{N}_i^{oo}} q_{ij}^{oo}} \, \hat{\bx}_{ij}^{oo}, \\
    &p_i^{oo'} = {\bw_{oo'}^\T}[\mathrm{ReLU}(\bW^{oo'}_{s} \, \bs^{oo})\circ \mathrm{ReLU}(\bW^{oo'}_{x}\,\tilde{\bx}_{i}^{oo})], \\
    &{p_i^{oo} = \mathbf{softmax}(p_i^{oo'}), \quad i = 1, \dots, N,}
    \end{aligned}}
\end{equation}
where $\tilde{\bx}_i^{oo}$ is considered as the question-conditioned {\bf oo} edge feature \wrt object node $o_i$.
We compute $\bp^{ot}$, $\bp^{tt}$ and $\bp^{to}$ using the same above equations, 
but with individual initial edge features, question representations and transformation parameters. 

\noindent{\bf Weighting Module}
The above graph attention modules output three attention weights for each object and text node, via the corresponding self-attended question part as the guidance. For each object node $o_i$, we have $p_i^{o}$, $p_i^{oo}$ and $p_i^{ot}$.
Similarly, for each text node $t_i$, we have $p_i^{t}$, $p_i^{tt}$ and $p_i^{to}$. Now we combine them together with the question self-attention weights. 
For each object node, the final weight score is calculated as a weighted sum of three parts: 
\begin{gather}
{
 \alpha^o_i = w^{o}p_i^{o} + w^{oo}p_i^{oo} + w^{ot}p_i^{ot}, \, i=1,\dots,N,
}
\end{gather}
where $w^{o,oo,ot}$ are obtained in Section \ref{sec:lan}. 
Similarly, the final weight for each text node is:
\begin{gather}
{
 \alpha^t_i = w^{t}p_i^{t} \!+\! w^{tt}p_i^{tt} \!+\! w^{to}p_i^{to}, \, i=N\!+\!1,\dots,N\!+\!M.
}
\end{gather}
Note that $\sum_{i=1}^{N}{\alpha^o_i} = 1$, as we have $w^o + w^{oo} + w^{ot} = 1$, $\sum_{i=1}^{N}{p^o_i} = 1$, $\sum_{i=1}^{N}{p^{oo}_i} = 1$ and $\sum_{i=1}^{N}{p^{ot}_i} = 1$. Likewise, we also have $\sum_{i=N+1}^{N+M}{\alpha^t_i} = 1$. 
The weights $\{ \alpha^o_i \}_{i=1}^{N}$ and $\{ \alpha^t_i \}_{i=N+1}^{N+M}$ actually measure the relevance between object/text nodes and the question, and are used to generate question-conditioned object and text features:

\begin{equation}\label{fobj}
{
 	\bg_{obj}= \textstyle{\sum_{i=1}^{N}}\alpha^o_i\cdot\hat{\bx}_i^{obj}, \quad
 	\bg_{text}=\sum_{i=N+1}^{N+M}\alpha^t_i\cdot\hat{\bx}_i^{text}.  
}
\end{equation}

\subsection{Global-Local Attentional Answering Module}
\label{sec:ans_mod}

Inspired by the iterative answer prediction mechanism in M4C~\cite{hu2019iterative}, we introduce it into our SMA by modifying the input of first decoding step as our global-local attentional answering module (in Figure \ref{fig:answerModule}). The input of first decoding step of M4C is a special token $\mathsf{<begin>}$, while we replace it with the the summarized global features of question, object and text, together with local OCR embeddings ($\bg_{obj}$ and $\bg_{text}$), which include comprehensive information of our question conditioned graph. 

The global graph features $\bg_{obj}$ and $\bg_{text}$ are generated by fusing global features of our question conditioned graph. Specifically, object-related and text-related question features are concatenated together firstly:
\begin{equation}
{
	\begin{aligned}
    \bar{\bs}^{o} = [\bs^{o}; \bs^{oo}; \bs^{ot}], \quad 
    \bar{\bs}^{t} = [\bs^{t}; \bs^{tt}; \bs^{to}].
    \end{aligned}
}
\end{equation}
$\bar{\bs}^{o}$, $\bar{\bs}^{t}$, $\bg_{obj}$, $\bg_{text}$ are forwarded into transformer layers along with local OCR node embeddings and updated as $\tilde{\bs}^{o}$, $\tilde{\bs}^{t}$, $\tilde{\bg}_{obj}$, $\tilde{\bg}_{text}$. Here, OCR embeddings have already been updated by graph attention module, \ie, they are $\{\alpha^t_i\cdot\hat{\bx}_i^{text}\}_{i=N+1}^{N+M}$, slightly different than Equation~\ref{fobj}. During transformer updation process, these global features and local OCR features can freely attend to each other.

Then we fuse updated features $\tilde{\bg}_{obj}$ and $\tilde{\bg}_{text}$ with their respective question representations as follows: 
\begin{equation}
{
	\begin{aligned}
    \bg_{OBJ}  &= \tilde{\bg}_{obj}  \circ \tilde{\bs}^{o}, \quad
    \bg_{TEXT} &= \tilde{\bg}_{text} \circ \tilde{\bs}^{t}.
    \end{aligned}
}
\end{equation}
The equation for predicting the answer probabilities in the first timestep $\bp^{1}_{ans}$ can be written as:
\begin{gather}
\bp^{1}_{ans} = \mathrm{f}_{pred}(\bW_{g}[\bg_{OBJ}; \bg_{TEXT}]), 
\end{gather}
where $\bW_{g}$ is a linear transformation and $\mathrm{f}_{pred}$ is a two-branch scoring function, which tackles the dilemma that answers in TextVQA task can be dynamic texts changing in different questions.
The rest timesteps' input and the answer space setting are same as which in M4C.

\noindent{\bf Training Loss}
\label{sec:training_loss}
Considering that the answer may come from two sources, we use multi-label binary cross-entropy (bce) loss:
\begin{gather}
{
    \begin{split}
    pred &= \frac{1}{1+\exp{(-y_{pred})}}, \\
    \mathcal{L}_{bce} &= -y_{gt} \mathrm{log}(pred)-(1-y_{gt}) \mathrm{log}(1-pred),
    \end{split}
}
\end{gather}
where $y_{pred}$ is prediction and $y_{gt}$ is ground-truth target.

\section{Experiments}
We evaluate our model on two challenging TextVQA benchmarks, including TextVQA~\cite{TexVQA} and all three tasks of ST-VQA~\cite{STVQA}, and achieve SoTA performance on TextVQA and the first two tasks of ST-VQA.
In our experiments, we find that the accuracy of OCR may limit the reasoning ability of the model, so
we manually labelled all the texts appeared in the TextVQA dataset, \ie, we provide the ground-truth of the OCR part, so that research community can fully study the reasoning ability of Text-VQA models, exempt from considering the effect of text recognition.

\makeatletter\def\@captype{table}\makeatother
{
\begin{center}
	\begin{table}[t]
	\centering
	\scalebox{1}{
	\setlength{\tabcolsep}{3mm}{
		\begin{tabular}{llccc}
        \toprule[1pt]
        \multicolumn{1}{c}{$\#$} & Method          & \begin{tabular}[c]{@{}c@{}}OCR\\ system\end{tabular}  & \multicolumn{1}{c}{\begin{tabular}[c]{@{}c@{}}Output\\ module\end{tabular}} & \begin{tabular}[c]{@{}c@{}}Accu.\\ on val\end{tabular}  \\ \hline \hline
        $1$ & Baseline       & Rosetta-ml  & classifier & $29.16$                                                                                                       \\
        $2$ & Baseline+oo     & Rosetta-ml   & classifier & $29.34$                                                                                                         \\
        $3$ & Baseline+ot     & Rosetta-ml   & classifier & $29.58$                                                                                                        \\
        $4$ & Baseline+tt     & Rosetta-ml  & classifier & $29.73$                                                                                                        \\
        $5$ & Baseline+to     & Rosetta-ml   & classifier & $30.14$                                                                                                         \\
        $6$ & Baseline+\textbf{all}  & Rosetta-ml   & classifier   & $30.26$    \\
        \bottomrule[1pt]
        \end{tabular}
	}
	}
	\caption{\label{tab:ablationStudyOnRosettaOCR}Ablation study on key components of question conditioned graph attention module on TextVQA dataset. 
	As mentioned in Section \ref{sec:lan_graph_attn}, there are four kinds of edges (relations) in our graph, which are \textbf{oo}, \textbf{ot}, \textbf{tt} and \textbf{to} edges. Stripping the question conditioned graph attention module of these four relations yields a baseline. 
	Individually, we add each of the four edge attentions into the baseline and evaluate the corresponding accuracy.
	}
	\end{table}
\end{center}}

\begin{table*}[htbp]
	\begin{center}
	\begin{tabular}{llcclccc}
    \toprule[1pt]
    \multicolumn{1}{c}{$\#$} & Method         & \begin{tabular}[c]{@{}c@{}}Question enc.\\ pretraining\end{tabular} & \begin{tabular}[c]{@{}c@{}}OCR\\ system\end{tabular} & \begin{tabular}[c]{@{}l@{}}OCR token\\ representation\end{tabular}  & \multicolumn{1}{c}{\begin{tabular}[c]{@{}c@{}}Output\\ module\end{tabular}} & \begin{tabular}[c]{@{}c@{}}Accu.\\ on val\end{tabular} & \begin{tabular}[c]{@{}c@{}}Accu.\\ on test\end{tabular} \\ \hline \hline
    $1$ & LoRRA~\cite{TexVQA}   & GloVe & Rosetta-ml & FastText   & classifier & $26.56$   & $27.63$     \\
    $2$ & LoRRA   & GloVe & Rosetta-en & FastText   & classifier & $29.35$   & -     \\
    $3$ & LoRRA   & GloVe & SBD-Trans & FastText   & classifier & $29.73$   & -     \\
    $4$ & DCD ZJU (ensemble)~\cite{DCDZJU}   & - & -   & - & - & $31.48$   & $31.44$     \\
    $5$ & MSFT VTI~\cite{MSFTVTI}   & - & - & -  & -  & $32.92$   & $32.46$     \\
    $6$ & M4C~\cite{hu2019iterative}   & BERT & Rosetta-ml & Multi-feats   & decoder & $37.06$   & -  \\
    $7$ & M4C   & BERT & Rosetta-en & Multi-feats   & decoder & $39.40$   & $39.01$  \\ 
    $8$ & M4C   & BERT & SBD-Trans & Multi-feats   & decoder & $40.24$   & -  \\
    $9$ & MM-GNN~\cite{gao2020multi}   & - & Rosetta-ml & -   & classifier & $31.44$   & $31.10$  \\
    $10$ & LaAP-Net~\cite{han2020finding}   & BERT & Rosetta-en & Multi-feats   & decoder & $40.68$   & $40.54$  \\
    $11$ & SA-M4C~\cite{sam4c}   & BERT & Google-OCR & Multi-feats   & decoder & $\textbf{45.40}$   & $44.60$  \\
    $12$ & PixelM4C~\cite{singh2021textocr}   & BERT & MTS v3 & Multi-feats   & decoder & $42.12$   & -  \\
    $13$ & TAP$^*$ w/o extra data~\cite{yang2020tap}   & BERT & Microsoft-OCR & Multi-feats   & decoder & $49.91$   & $49.71$  \\
    $14$ & TAP$^*$~\cite{yang2020tap}   & BERT & Microsoft-OCR & Multi-feats   & decoder & $54.71$   & $53.97$  \\
    \hline
    $15$ & \sexyname w/o dec. (Model 6 in Tab.\ref{tab:ablationStudyOnRosettaOCR}) & GloVe & Rosetta-ml  & FastText & classifier & $30.26$  & -  \\
    $16$ & \sexyname w/o dec. & GloVe & Rosetta-en  & FastText & classifier & $32.28$  & -  \\
    $17$ & \sexyname w/o dec. & GloVe & Rosetta-en  & Multi-feats  & classifier & $35.03$  & -  \\
    $18$ & \sexyname with M4C dec.  & GloVe & Rosetta-en  & Multi-feats  & decoder & $38.91$  & -  \\
    $19$ & \sexyname  & GloVe & Rosetta-en  & Multi-feats  & decoder & $39.36$  & -  \\
    $20$ & \sexyname & BERT & Rosetta-en & Multi-feats         & decoder  & $40.24$ & $40.21$ 
    \\
    $21$ & \sexyname & BERT & Rosetta-ml & Multi-feats + RecogCNN         & decoder  & $37.74$ & $-$  \\
    $22$ & \sexyname & BERT & Rosetta-en & Multi-feats + RecogCNN         & decoder  & $40.39$ & $40.86$  \\
    $23$ & \sexyname & BERT & SBD-Trans & Multi-feats + RecogCNN         & decoder  & $43.74$ & $44.29$  \\
    $24$ & \sexyname with ST-VQA Pre-training & BERT & SBD-Trans & Multi-feats + RecogCNN         & decoder  & $44.58$ & $\textbf{45.51}$  \\
    \bottomrule[1pt]
    \end{tabular}
	\end{center}
	\caption{\label{tab:TextVQA} 
	More ablation models and comparision to previous work. Compared to our SMA, TAP models need an extra pre-training stage with a group of pre-training tasks. The TAPs in lines 13 and 14 are pretrained with the training set of TextVQA dataset and an additional large-scale training data, respectively. }
	\label{my-label}
\end{table*}

\subsection{Implementation Details}
Following M4C~\cite{hu2019iterative}, the objects' and OCRs' region based appearance features are extracted from the fc6 layer which immediately follows the RoI-Pooling layer of a Faster R-CNN~\cite{ren2015faster} model. The model is pretrained on Visual Genome~\cite{krishna2017visual} and then fine-tuned fc7 layer on TextVQA~\cite{TexVQA}. 
The maximum number of object regions $N = 36$. 
For text nodes, we use four independent OCR methods to recognize word strings.
\begin{enumerate}
    \item Rosetta-ml OCR. Multi-language version of Rosetta system~\cite{borisyuk2018rosetta}.
    \item Rosetta-en OCR. English-only version of Rosetta system.
    \item SBD-Trans OCR. An SoTA Sequential-free Box Discretization (SBD) model~\cite{SBD} is for scene text detection and a robust transformer based network~\cite{textTrans} is for word recognition. Their training process will be detailed in Appendix F.
    \item Ground-Truth OCR. Its collecting process will be detailed in Section~\ref{sec:ocr_perform}.
\end{enumerate}

We recognize at most $M = 50$ OCR tokens in an image and generate rich OCR representations based on them. If any of the above is below maximum, we apply zero padding to the rest. 
We set the maximum length of questions to $T = 20$ and encode them as $768$D feature sequences by the first three layers of a pretrained BERT~\cite{BERT}, 
whose parameters are further fine-tuned during training. Our answering module uses $4$ layers of transformers with $12$ attention heads. The other hyper-parameters are the same with BERT-BASE~\cite{BERT}. The maximum number of decoding steps is set to $L = 12$.

We implement all the models in PyTorch and experiment on $4$ NVIDIA GeForce 1080Ti GPUs with a batch size of $96$.
The learning rate is set to $1e-4$ for all layers except for the three-layer BERT used for question encoding and the fc7 layer used for region feature encoding, which have a learning rate of $1e-5$. 
We multiply the learning rate by $0.1$ at the $14000$ and $19000$ iterations and the optimizer is Adam. 
At every $1000$ iterations we compute a VQA accuracy metric~\cite{goyal2017making} on the validation set, based on all of which the best performing model is selected. 
To gracefully capture errors in text recognition, the ST-VQA dataset~\cite{STVQA} adopts Average Normalized Levenshtein Similarity (ANLS) as its official evaluation metric.

\subsection{Results and Analysis on TextVQA}
The TextVQA dataset~\cite{TexVQA} 
samples $28,408$ images from OpenImages dataset~\cite{krasin2017openimages}.
The questions are divided into train, validation and test splits with size $34,602$, $5,000$, and $5,734$ respectively, and each question-image pair has $10$ human-provided ground truth answers. 

\subsubsection{Ablation Studies}
\noindent{\bf Ablations on Relationship Attentions.}
\label{sec:ablationRosetta}
We conduct an ablation study to investigate the key components of the proposed Question Conditioned Graph Attention Module, \ie, the four types ({\bf oo}, {\bf ot}, {\bf tt}, {\bf to}) of relationship attentions.
In order to focus on the reasoning ability, we evaluate it without rich OCR representation and iterative answering module.
The tested architecture variations and their results are shown in Table~\ref{tab:ablationStudyOnRosettaOCR}. 
In {\bf Baseline} model, the Role-Aware Graph in Figure~\ref{fig:fullmodel} has isolate OCR and object nodes, dispensing with interactions between nodes and matching question features.
Despite the similarity, {\bf to} and {\bf ot} are intrinsically different. Text-obj edges are text-centered, with neighbouring object nodes as context to update text features, whereas obj-text relations are object-centered, with neighbouring text nodes as context to update object features. The two types of edges have different roles in feature updating. 
The experimental results reveal that each of the four modeled relations has improved the accuracy. 
In particular, the {\bf to} relation attention leads to the largest improvement among all kinds. 
It is consistent with the observation that annotators tend to refer to a specific text by describing the object where the text is printed on.  
Overall, the relations whose origins are text ({\bf to} and {\bf tt}) are more important than those for object ({\bf oo} and {\bf ot}),
which validates the key role of text in this text VQA task. The Baseline+all model combines all the relationships and leads to the best performance.

\noindent{\bf Ablations on Answering Modules.}
From model $17$ and $19$ of Table~\ref{tab:TextVQA}, we can find that our proposed generative answering module surpasses the discriminative classifier-based answering module by a large margin ($4.33\%$ in validation accuracy). We also verify the modification of our answering model by comparing it with the original structure for answer generation proposed by M4C~\cite{hu2019iterative} in $18$ and $19$ of Table~\ref{tab:TextVQA}, which shows that our modification can bring nearly $0.5\%$ of accuracy improvement.

\begin{figure*}[t]
	\centering
	\begin{center}
		\includegraphics[width=\textwidth]{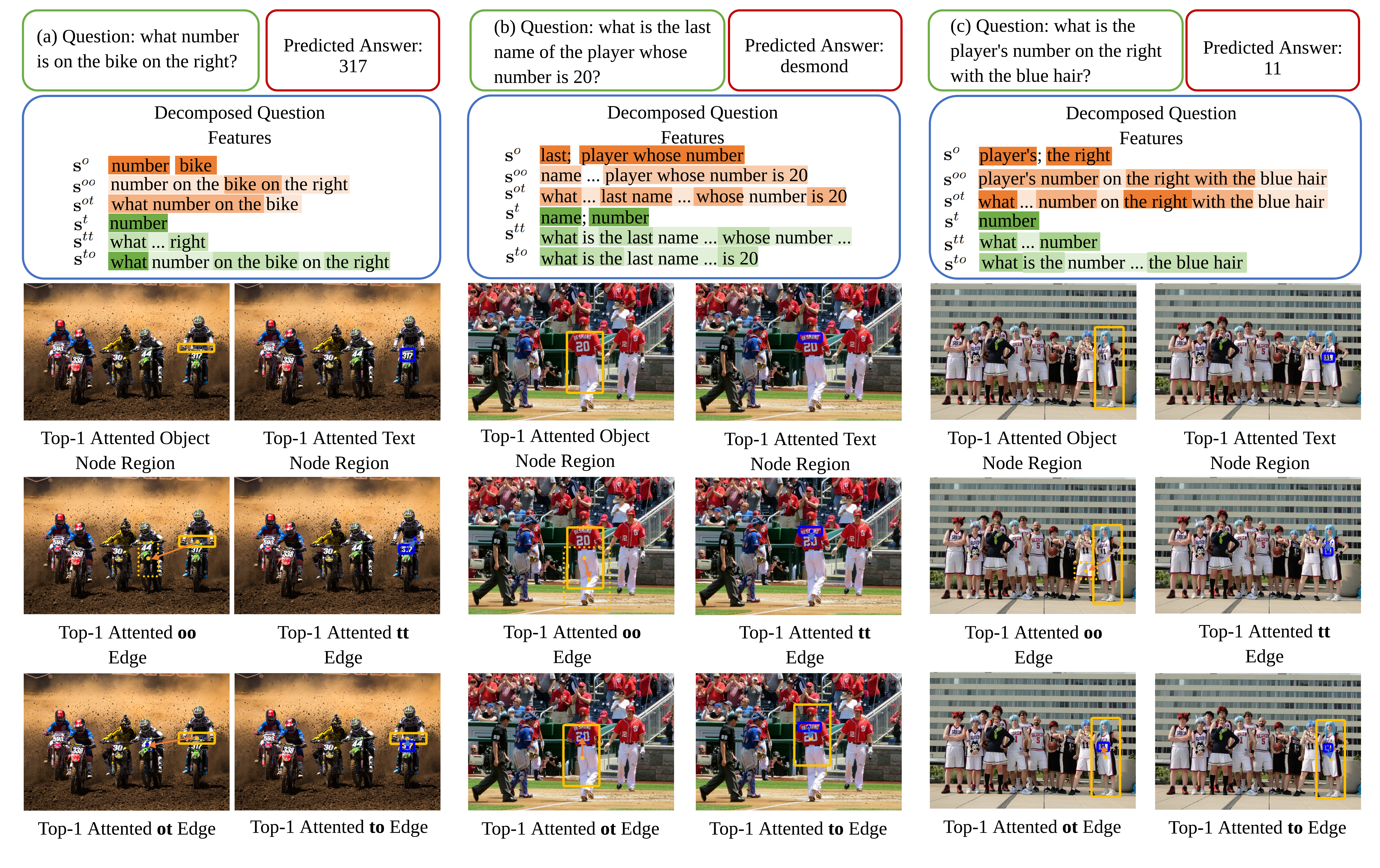}
		\caption{{\bf Edge attention and decomposed question attention visualization for \sexyname.} Three representative examples that require relationship reasoning for question answering are presented, which demand different kinds of edge relations. For instance, {\bf to} represents the relation whose former node is {\bf t}ext and latter one is {\bf o}bject. For each example we highlight nodes or edges with the highest attention weights, wherein nodes are represented by boxes and edges are displayed by arrows pointing from former node to latter one. 
	    For boxes/nodes, yellow ones are for object and blue ones are for text. Solid ones are those with the highest attention weights whereas dashed ones are normal. 
	    For decomposed question attention, the darker highlighted text area has a higher attention weight. 
	    All of them are predicted by \sexyname with Ground-Truth OCR. It can be seen that question attention module successfully figures out desired relations.}
		\label{fig:visualization}
	\end{center}
\end{figure*}

\noindent{\bf Ablations on Features for Question and OCR.}
From model $16$ to $17$, there are nearly $3\%$ improvements gained by replacing the FastText features with Multi-features of OCR tokens. The Multi-feature is a feature package including FastText, Faster R-CNN, PHOC and BBox features proposed in M4C \cite{hu2019iterative}.
The Glove and BERT features are evaluated for encoding questions, and the latter outperforms by $0.88\%$ in validation accuracy (see models $19$ and $20$ in Table~\ref{tab:TextVQA}).
By comparing models $20$ and $22$ in Table~\ref{tab:TextVQA}, we can see a further improvement of $0.15\%$ on validation set and $0.65\%$ on test split by adding the RecogCNN feature for OCRs.
This validates that the RecogCNN feature is complementary to the Multi-Feats. 
Note that RecogCNN is trained on a text recognition task while Faster R-CNN is trained for general object detection.
FastText and PHOC are extracted from the recognized OCR character sequences, but RecogCNN is extracted from text visual patches.

\noindent{\bf Visualization.} For each type of relations, we visualize those with the highest attention weights and their corresponding decomposed question attention, in order to explore their contributions in answer prediction and give better insights in explaining our model (see Figure \ref{fig:visualization}). 
In the first example, there are several bikes among which the question asks about the right one. Locating the requested bike needs {\bf oo} relationship reasoning. Another relationship {\bf to} is also in need as the number on the bike is exactly what we have to figure out. In the second example, we need {\bf ot} relationship to locate the player whose number is $20$. The {\bf to} relationship is employed then to reason about the last name of this player.
Similarly, in the last example, two different {\bf oo} relationships are extracted to pinpoint the location of the player on the right and with blue hair. Then {\bf ot} relationship is used to get the player's number.
All the examples validate the relationship reasoning ability of our model.

\subsubsection{Comparison to Previous Work.} 
We compare our method to previous methods and achieve surpassing results. 
Using the same question, object and OCR features, our single model is $1.20\%$ better than M4C (model $20$ VS. $7$) on the test set. Our final model $24$, by applying an advanced OCR system SBD-Trans (more details can be found in Section~\ref{sec:ocr_perform}), achieves $44.29\%$ on test split. 
With ST-VQA pre-training, we further achieve $45.51\%$ on test split, which is the current state-of-the-art (except the pre-training based TAP~\cite{yang2020tap}), outperforming the previous best model M4C by a large margin, \ie $6.5\%$ and won the TextVQA Challenge 2020 (more details please refer to Appendix E).

\makeatletter\def\@captype{table}\makeatother
{
\begin{center}
	\begin{table}[t]
	\centering
	\scalebox{1}{
	\setlength{\tabcolsep}{3mm}{
		\begin{tabular}{cccc}
        \toprule[1pt]
            \multirow{2}{*}{\#} & \multirow{2}{*}{Methods} & \multicolumn{2}{c}{Val Accuracy (\%)} \\ \cline{3-4} 
                                &                          & w/o GT OCR           & w GT OCR            \\ \hline\hline
            1                   & LoRRA~\cite{TexVQA}                    & $29.35$             & $35.07$             \\
            2                   & M4C~\cite{hu2019iterative}                      & $39.40$            & $47.91$             \\
            3                   & SMA (Ours)               & $40.39$             & $50.07$             \\
            4                   & OCR UB                   & $44.98$                 & $68.81$             \\
            5                   & Human                    & $85.01$             & $85.01$             \\ \bottomrule[1pt]
        \end{tabular}
	}
	}
	\caption{\label{tab:ablationOnGTOCR}Evaluation with GT OCR on three models. After replacing Rosetta-en OCR with GT OCR, all of them improve by a large margin. An OCR UB metric is also provided for reference, which measures the upper bound vqa accuracy of a certain OCR system.
	}
	\end{table}
\end{center}}

\subsection{Effect of OCR Performance on TextVQA}
\label{sec:ocr_perform}
However, during the experiments, we find that sometimes even our model can attend to the right OCR regions that are related to the question, we still can not answer it correctly. This mainly because the OCR tokens are failed to be recognized by the OCR model. This illustrates that the performance of OCR, \ie, the reading ability greatly influences the reasoning ability of model. See Appendix A for examples. To fully investigate this problem and set an upper bound for our \sexyname model (and previous models), we provide the human annotation of all the OCR tokens in the images of TextVQA. The annotation are available at https://github.com/ChenyuGAO-CS/SMA

\subsubsection{Human Annotated Ground-Truth OCR. }
We provide a ground-truth OCR annotation of the TextVQA train and validation sets, because it provides a fair test base for researchers to focus on the text-visual reasoning part without tuning the OCR model additionally. More detail please refer to Appendix B.


\subsubsection{Results with Ground-Truth OCR.}
We evaluate the performance of LoRRA, M4C and \sexyname, using the ground-truth OCR. The results are shown in Table \ref{tab:ablationOnGTOCR}.
Both of them improve by a large margin: LoRRA goes up from $29.35\%$ to $35.07\%$ and M4C increases from $39.40\%$ to $47.91\%$ while \sexyname shoots up to $50.07\%$ from $40.39\%$ on the validation set, 
by replacing Rosetta-en results with ground truth. 
The larger increase in accuracy ($9.68\%$ vs $5.72\%$ and $8.51\%$) demonstrates better reasoning and answering ability of our model.
OCR UB is the upper bound accuracy one can get if the answer can be build using single or multiple token(s) from the OCR source inspired by~\cite{hu2019iterative}. 
With GT OCR, the upper bound can be promoted to $68.81\%$ on validation. 
However, there's still a large gap between human performance and \sexyname, which has great potential for us to unlock.


\subsubsection{Model Performance with Different OCRs}
\begin{figure}[t!]
	\begin{center}
		\includegraphics[width=0.48\textwidth]{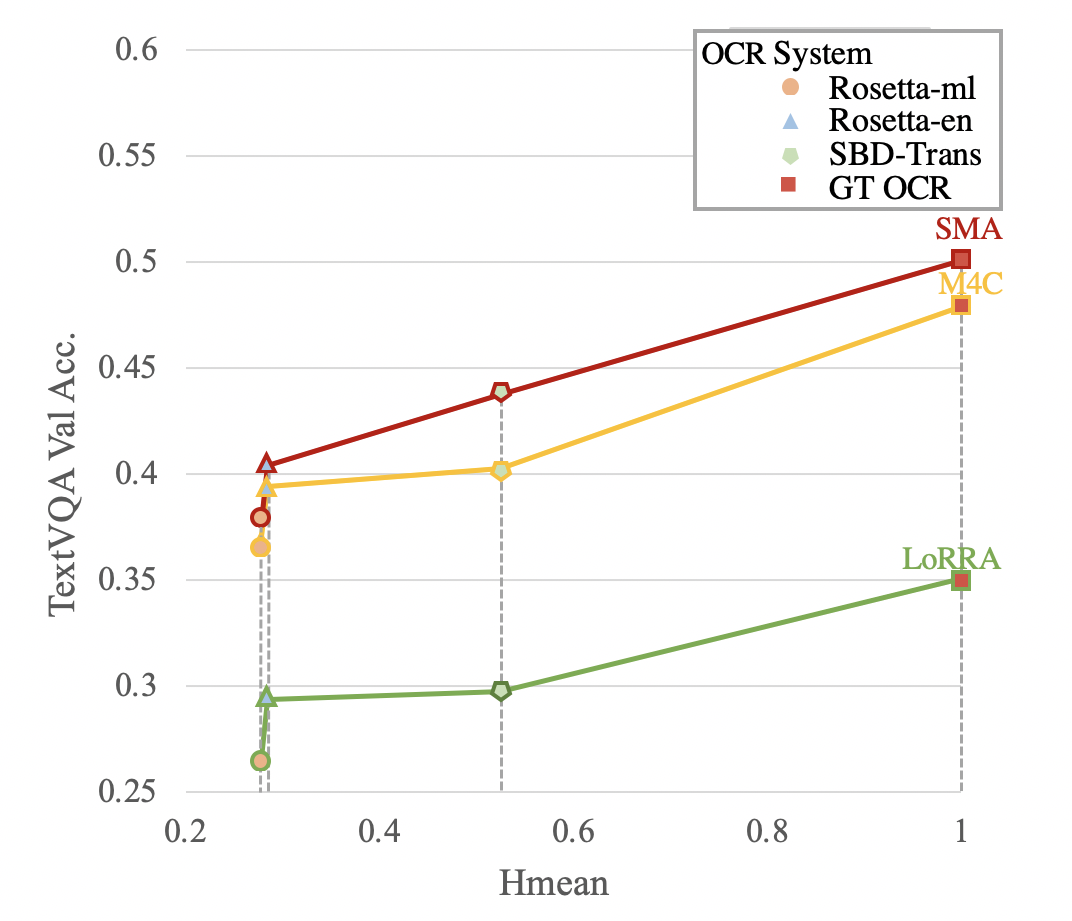}
	\end{center}
	\caption{Performance of LoRRA, M4C and SMA with different OCR systems on the TextVQA dataset. Hmean measures the ability of OCR models. The x-axis shows the Hmean value of four
	gradually increased OCRs (Rosetta-ml, Rosetta-en, SBD-Trans and Ground-Truth). The y-axis presents the val accuracy of reasoning models.}
	\label{fig:AccOCR}
\end{figure}

We compare the accuracy of different models with 4 types of OCR systems (which are Rosetta-ml, Rosetta-en and SBD-Trans and Ground-Truth) in Figure~\ref{fig:AccOCR}. It's clear that, with more accurate OCR system, the accuracy of LoRRA, M4C and SMA are all improved. Our proposed \sexyname model performs best. 

For reference, we also provide the OCR accuracy of Rosetta-ml, Rosetta-en and SBD-Trans respectively on the basis of manually labelled ground truth, as shown in Table~\ref{tab:OCR_Accuracy}. We find Rosetta-ml fails to show promising results and Rosetta-en performs slightly better, while SBD-Trans performs much better, especially on ‘Recall’ and ‘Hmean‘.

\makeatletter\def\@captype{table}\makeatother
{
\begin{center}
\begin{table}[t]
\centering
    \setlength{\tabcolsep}{4mm}{
	\begin{tabular}{lccc}
		\toprule[1pt]
		\multicolumn{1}{l}{Methods} & Precision & Recall   & Hmean  \\ \hline \hline
		Rosetta-ml                      & $0.4789$  & $0.1959$ & $0.2781$ \\ 
		Rosetta-en                      & $0.5106$  & $0.1966$ & $0.2839$ \\ 
		SBD-Trans               & $0.5958$  & $0.4683$ & $0.5244$ \\ \hline
		\bottomrule[1pt]
	\end{tabular}
	}
	\caption{An OCR result is considered as a match if its bounding box overlaps with corresponding ground-truth one by over $50\%$ of the total area and the tokens given are the same. It can be clearly seen that SBD-Trans OCR is more accurate than Rosetta-ml OCR and Rosetta-en OCR. }
\label{tab:OCR_Accuracy}
\end{table}
\end{center}}

\makeatletter\def\@captype{table}\makeatother

\begin{center}
	\begin{table}[t]
	\centering
	\scalebox{1}{
	\setlength{\tabcolsep}{3mm}{
		\begin{tabular}{clccc}
        \toprule[1pt]
        $\#$ & \multicolumn{1}{c}{Method} & \begin{tabular}[c]{@{}c@{}}Task 1\\ ANLS\end{tabular} & \begin{tabular}[c]{@{}c@{}}Task 2\\ ANLS\end{tabular} & \begin{tabular}[c]{@{}c@{}}Task 3\\ ANLS\end{tabular} \\ \hline \hline
        1   & SAN+STR~\cite{STVQA}   & $0.135$    & $0.135$      & $0.135$           \\
        2   & VTA~\cite{biten2019icdar}  & $0.506$    & $0.279$    & $0.282$                \\
        3   & M4C~\cite{hu2019iterative}  & $-$    & $-$        & $0.462$        \\
        4   & MM-GNN~\cite{gao2020multi}  & $-$    & $0.203$        & $0.207$ \\
        5   & LaAP-Net~\cite{han2020finding}  & $-$    & $-$        & $0.485$        \\
        6   & SA-M4C~\cite{sam4c}  & $-$    & $-$        & $\textbf{0.504}$        \\
        7   & TAP$^*$ w/o extra data~\cite{yang2020tap}  & $-$    & $-$        & $0.543$        \\
        8   & TAP$^*$~\cite{yang2020tap}  & $-$    & $-$        & $0.597$        \\
        9   & \sexyname (Ours)   & $\textbf{0.508}$   & $\textbf{0.483}$     & $0.486$    \\ \bottomrule[1pt]
        \end{tabular}
	}
	}
	\caption{\label{tab:stVQA}Evaluation on ST-VQA dataset. Compared to our SMA, TAP models need an extra pre-training stage with a group of pre-training tasks. The TAPs in lines 7 and 8 are pretrained with the training set of ST-VQA dataset and an additional large-scale training data, respectively. }
	\end{table}
\end{center}

\subsection{Evaluation on the ST-VQA dataset}
The ST-VQA dataset~\cite{STVQA} comprises of $23,038$ images with $31,791$ question-answer pairs. There are three VQA tasks, namely strongly contextualised, weakly contextualised and open vocabulary. For the strongly contextualised task, the authors provide a $100$-word dictionary per image; in the weakly contextualised task, the authors provide a single dictionary of $30,000$ words for all images and for the open dictionary task, no candidate answers are provided. 
As the ST-VQA dataset does not have an official split for training and validation, we follow M4C~\cite{hu2019iterative} to randomly select $17,028$ images as our training set and use the remaining $1,893$ images as our validation set.

For the first task, a single-step version of our model (\sexyname w/o dec.) is used, while in the second and third task we use the proposed full \sexyname model with different timesteps(3 for task 2 and 12 for task 3). 
Compared with methods on the leaderboard, we set new SoTA for the first two tasks and comparable result for the Task 3 (see Table~\ref{tab:stVQA}). 


\section{Conclusion}
We introduce Structured Multimodal Attentions (\sexyname), a novel model architecture for answering questions based on the texts in images, that sets new state-of-the-art performance on the TextVQA and ST-VQA dataset. 
\sexyname is composed of three key modules: a {\em Question Self-Attention Module} that guides a {\em Graph Attention Module} to learn the node and edge attention, and a final {\em Answering Module} which combines the attention weights and question-guided features of aforementioned Graph Attention Module to yield a reasonable answer iteratively. 
Alongside SMA model, we also conduct thorough experiments with several OCR systems and analyze how much they can influence overall performance.
A human-annotated ground-truth OCR set of TextVQA is also provided to set up the new upper bound and to help the community evaluate the real text-visual reasoning ability of different models, without suffering from poor OCR accuracy.


\ifCLASSOPTIONcaptionsoff
  \newpage
\fi



\bibliographystyle{IEEEtran}
\bibliography{egbib}
%


%

\vspace{-1.4cm}
\begin{IEEEbiography}[{\includegraphics[width=1in,height=1.25in,clip,keepaspectratio]{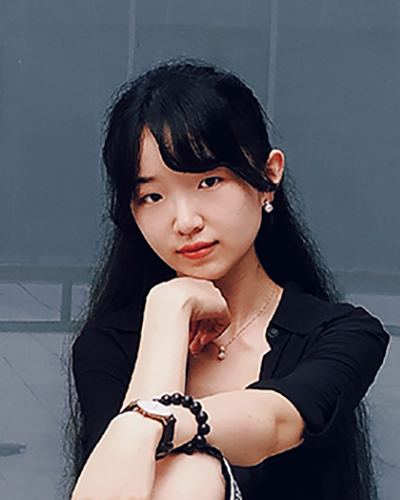}}]{Chenyu Gao} received the Bachelor degree in software engineering from the Northwestern Polytechnical University, China, in 2019. She is currently pursuing the MSc degree from School of Software Engineering, Northwestern Polytechnical University. Her research interests include computer vision, machine learning and artificial intelligence.
\end{IEEEbiography}

\vspace{-1.4cm}
\begin{IEEEbiography}[{\includegraphics[width=1in,height=1.25in,clip,keepaspectratio]{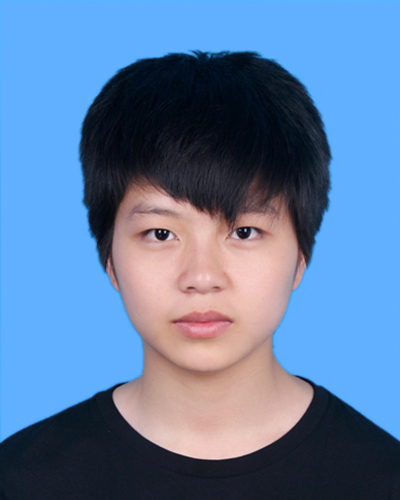}}]{Qi Zhu} received the B.E. degree in Computer Science from Northwestern Polytechnical University, China. Her research interest are computer vision and machine learning.
\end{IEEEbiography}

\vspace{-1.4cm}
\begin{IEEEbiography}[{\includegraphics[width=1in,height=1.25in,clip,keepaspectratio]{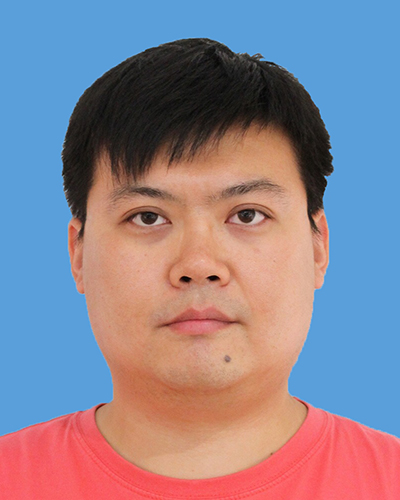}}]{Peng Wang} is a Professor at School of Computer Science, Nothwestern Polytechnical University, China. He was with School of Computer Science, the University of Adelaide for about four years. His research interests are computer vision, machine learning and artificial intelligence. He received a Bachelor in electrical engineering and automation, and a PhD in control science and engineering from Beihang University (China) in 2004 and 2011, respectively.
\end{IEEEbiography}

\vspace{-1.4cm}
\begin{IEEEbiography}[{\includegraphics[width=1in,height=1.25in,clip,keepaspectratio]{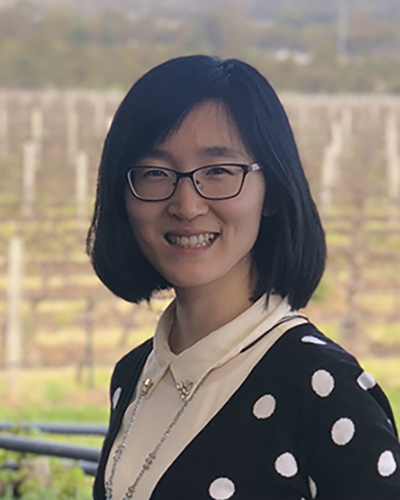}}]{Hui Li} received the PhD degree in computer science from the University of Adelaide (Australia). She is now a postdoctoral researcher at the Australian Centre for Robotic Vision (ACRV). Her research interests include deep learning, scene text recognition and visual question answering.
\end{IEEEbiography}

\vspace{-1.4cm}
\begin{IEEEbiography}[{\includegraphics[width=1in,height=1.25in,clip,keepaspectratio]{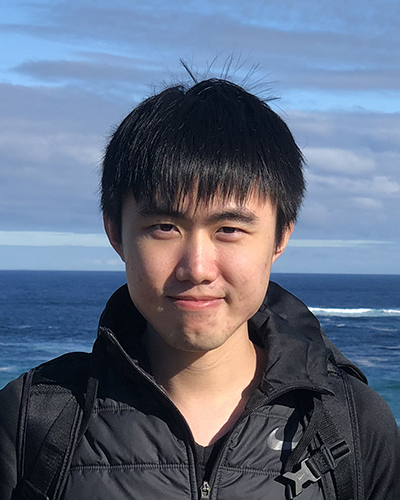}}]{YuLiang Liu} received the B.S. degree in electronic and information engineering from the South China University of Technology, China, in 2016, where he received the Ph.D. degree with Deep Learning and Vision Computing Lab (DLVC-Lab), under the supervision of Prof. Lianwen Jin. He is now a postdoc in University of Adelaide, under the supervision of Prof. Chunhua Shen. He is working on scene text understanding, handwritten character recognition, document analysis, and deep learning-based text detection and recognition.
\end{IEEEbiography}

\vspace{-1.4cm}
\begin{IEEEbiography}[{\includegraphics[width=1in,height=1.25in,clip,keepaspectratio]{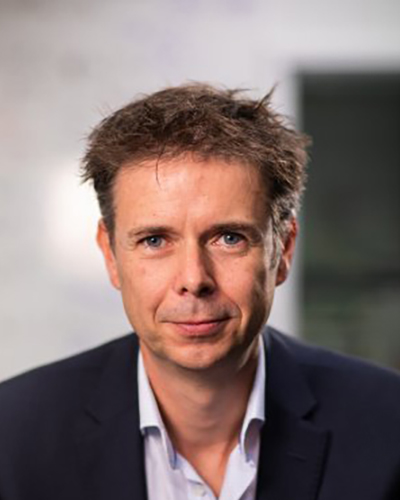}}]{Anton Van Den Hengel} is a Professor at the University of Adelaide and the founding Director of The Australian Centre for Visual Technologies (ACVT). He received a PhD in Computer Vision in 2000, a Master Degree in Computer Science in 1994, a Bachelor of Laws in 1993, and a Bachelor of Mathematical Science in 1991, all from The University of Adelaide.
\end{IEEEbiography}

\vspace{-1.4cm}
\begin{IEEEbiography}[{\includegraphics[width=1in,height=1.25in,clip,keepaspectratio]{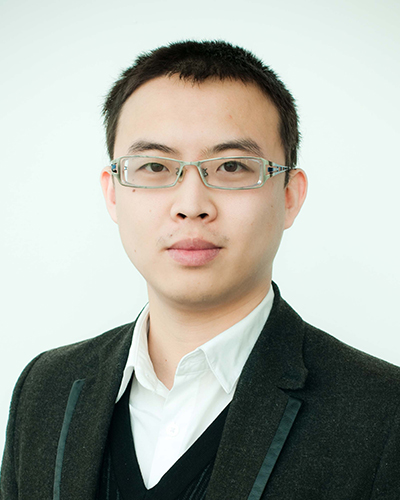}}]{Qi Wu} is  a  Senior Lecturer  (Assistant  Professor)  in  the University  of  Adelaide  and  he  is  an  Associate  Investigator in the Australia Centre for Robotic Vision (ACRV).  He  is  the  ARC  Discovery  Early  Career Researcher Award (DECRA) Fellow between 2019-2021. He obtained his PhD degree in 2015 and MSc degree in 2011, in Computer Science from the University  of  Bath,  United  Kingdom.  His  educational background  is  primarily  in  computer  science  and mathematics. He works on the Vision and Languagep roblems, including Image Captioning, Visual Ques-tion Answering, Visual Dialog etc. His work has been published in prestigious journals and conferences such as TPAMI, CVPR, ICCV, AAAI and ECCV. 
\end{IEEEbiography}







\onecolumn
\newpage

\setcounter{section}{0}
\renewcommand\thesection{\Alph{section}}
\newcommand{\xiaoer}{\fontsize{14pt}{14pt}\selectfont}
\begin{center}
\xiaoer\textit{\textbf{Appendix}}
\end{center}

\section{Effect of OCR Performance on TextVQA}
\begin{figure*}[b!]
	\centering
	\begin{center}
		\includegraphics[width=\textwidth]{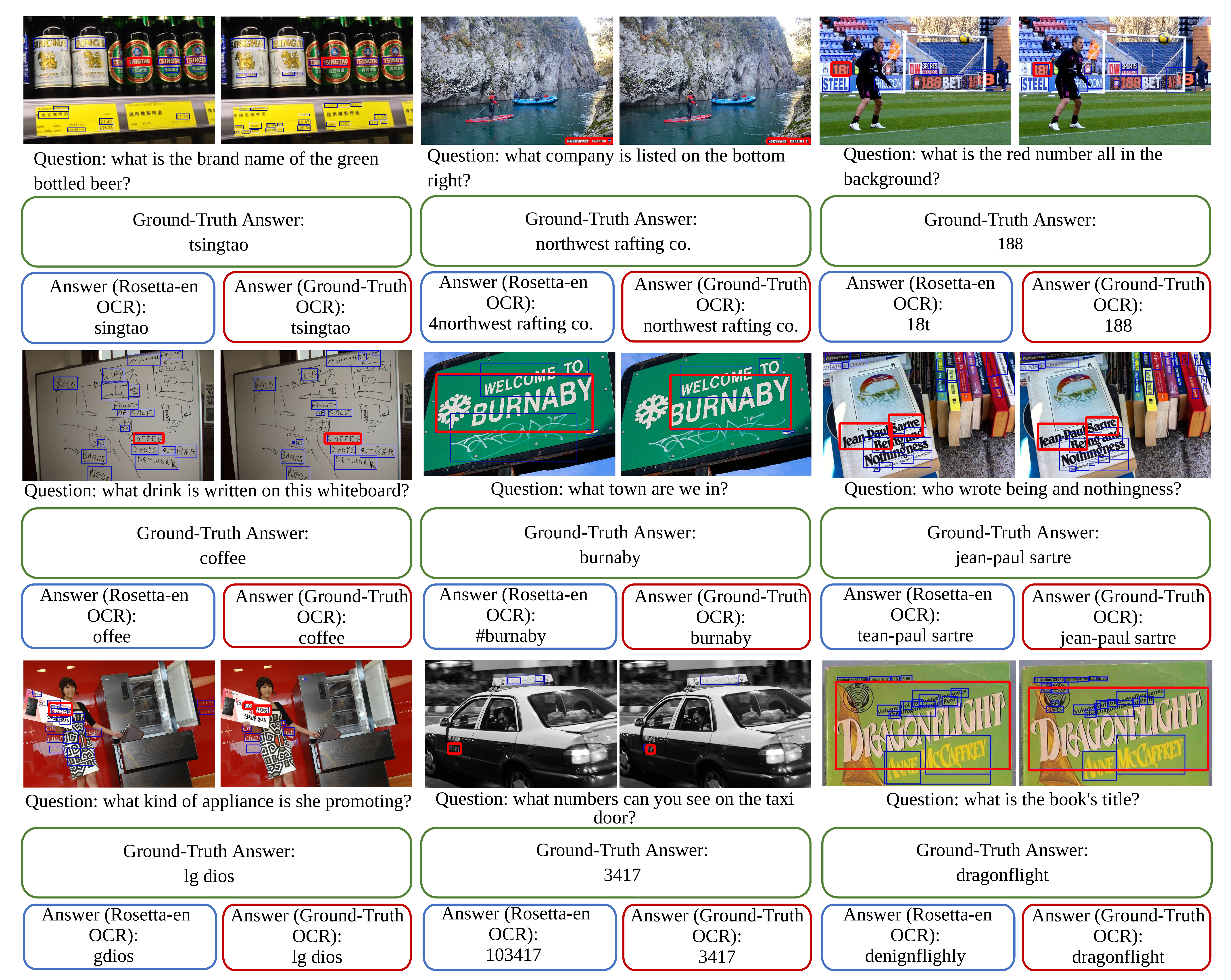}
		\caption{{\bf \sexyname's predictions for TextVQA examples.} We select multiple examples whose answers lie in OCR tokens from TextVQA. Green-dash boxes contain the most frequently occurring answer among 10 Ground-Truth answers for each question in TextVQA dataset. In the images, OCR annotations are in blue boxes, the OCRs who receive the highest score are in red boxes. Answers in blue-dash boxes and red-dash boxes are generated by \sexyname with Rosetta-en OCR and Ground-Truth OCR respectively. Even right reasoning process can not lead to a fully correct answer due to defective OCR results.}
		\label{fig:GT-example}
	\end{center}
\end{figure*}

This analysis is performed on the validation set of TextVQA. In Figure~\ref{fig:GT-example}, we visualize the prediction results of \sexyname with both Rosetta-en OCR and Ground-Truth OCR. 
We can find that, when the OCR area related to the question is correctly detected by the machine, \sexyname based on both Rosetta-en OCR and Ground-Truth OCR can attend to the most relevant OCR region in many cases. However, slight recognition error may still result in the wrong answer. 
If the OCR area related to the question cannot be detected by the machine at all, then the question has no chance of being answered correctly. 
This illustrates that the performance of OCR, \ie, the reading ability greatly influences the reasoning ability of model.

\section{Human Annotated Ground-Truth OCR}
The reason that we provide a ground-truth OCR annotation is that it will provide a fair test base for this area, so that researchers can focus on the text-visual reasoning part without tuning the OCR model.
We ask Amazon Mechanical Turk (AMT) workers to annotate all the texts appearing in the TextVQA dataset in order to completely peel off the impact of OCR and to investigate the real reasoning ability on a fair test base. 

To ease the labour for workers, We first produce a set of OCR results with the trained text detection and recognition model. Then the AMT workers check whether the bounding boxes and texts generated by the machine are correct or not, after which is a four-branch process:
first, if the bounding box is completely wrong, the workers are asked to delete it directly; 
the second case, if the text is detected correctly but recognised wrong, the right text is given by the worker;
thirdly, if the position of the bounding box is just not accurate enough, slight modifications are applied to it; 
finally, when a text is missed by machine, the workers draw the bounding box and provide label for the text region at the same time.

\section{Failure Cases of SMA.} 
\begin{figure*}[t]
	\centering
	\begin{center}
		\includegraphics[width=0.98\textwidth]{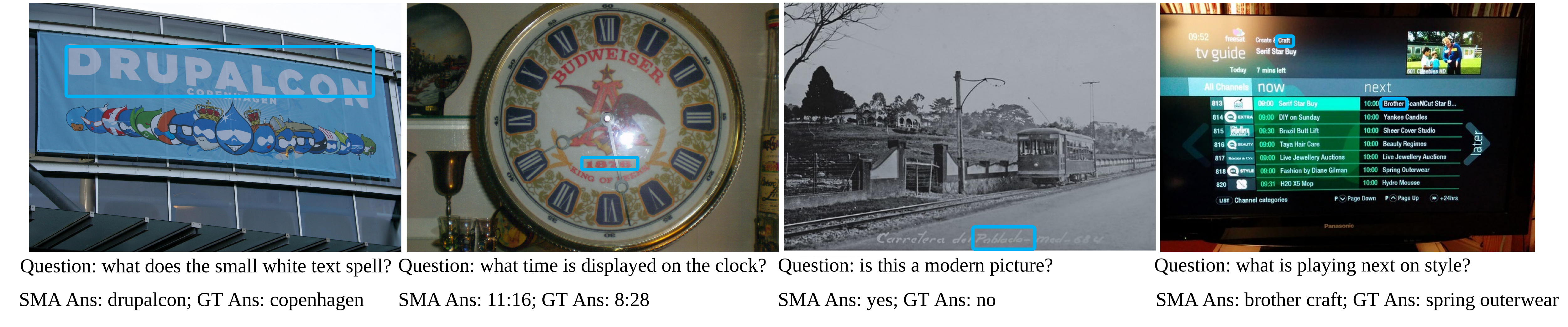}
		\caption{Typical failure cases of \sexyname on TextVQA validation set on the basis of Ground-Truth OCR. We highlight highest-score OCR with blue box. (1) SMA can not distinguish between ``small'' and ``large''. (2) The gap between clock hand angle and ``xx:xx'' formatted answer overwhelms reasoning model. (3) Commensense knowledge needed. (4) Expected answer is out of range of considered $50$ OCR tokens.}
		\label{fig:failure}
	\end{center}
\end{figure*}

We show and analyze some failure examples of our model in Figure~\ref{fig:failure}. 
In the first example, our model reasons correctly about the color - ``white'', however, fails at the size property - ``small''. Some commonsense knowledge about size is needed for this question.
The second example of asking about time is quite tricky due to the gap between clock hand angle and desired ``xx:xx'' formatted answer, which demands sophisticated reasoning ability from the model.
The third example is very hard for our model to answer, as it require background knowledge that is hard to grasp from current limited dataset. 
In the fourth image, as the desired answer lies out of range of the specified $50$ OCR tokens, there is no way to answer it right.

\section{TextVQA Challenge 2020}
In TextVQA Challenge 2020, it is not allowed to use ensemble models. 
The runner-up model SA-M4C~\cite{sam4c} ($44.80\%$ on test set) upgraded from vanilla transformers in M4C~\cite{hu2019iterative} to a spatially aware self-attention layer. Besides, several tricks such as better detector backbone, beam search decoding, two additional self-attention layers, better OCR system (Google OCR system) and jointly training with ST-VQA dataset~\cite{STVQA} are also used in their method. In contrast, we only used a single SMA model with SBD-Trans OCR system and ST-VQA dataset as additional training data, which achieved $\textbf{45.51\%}$ final test accuracy – a new SoTA on the TextVQA dataset.  

\section{SBD-Trans Training Data.}
The SBD model is pretrained on a $60k$ dataset, which consists of $30,000$ images from LSVT~\cite{sun2019icdar} training set, $10,000$ images from MLT 2019~\cite{nayef2019icdar2019} training set, $5,603$ images from ArT~\cite{chng2019icdar2019} (containing all the images of SCUT-CTW1500~\cite{liu2019curved} and Total-text~\cite{ch2017total,ch2020total}), and $14,859$ images selected from a bunch of datasets (RCTW-17~\cite{shi2017icdar2017}, ICDAR 2013~\cite{karatzas2013icdar}, ICDAR 2015~\cite{karatzas2015icdar}, MSRA-TD500~\cite{yao2012detecting}, COCOText~\cite{veit2016coco}, and USTB-SV1K~\cite{yin2015multi}). The model was finally finetuned on MLT 2019~\cite{nayef2019icdar2019} training set. 
The robust transformer based network is trained on the following datasets: IIIT 5K-Words~\cite{mishra2012scene}, Street View Text~\cite{wang2011end}, ICDAR 2013~\cite{karatzas2013icdar}, ICDAR 2015~\cite{karatzas2015icdar}, Street View Text Perspective~\cite{quy2013recognizing}, CUTE80~\cite{risnumawan2014robust} and ArT~\cite{chng2019icdar2019}.


\end{document}